\documentclass[10pt,journal,compsoc]{IEEEtran}

\usepackage{xcolor}
\usepackage[T1]{fontenc}
\usepackage{epsfig}
\usepackage{amsmath,bm}
\usepackage{amsthm,amssymb,amsfonts}
\usepackage{algorithm}
\usepackage{algorithmic}
\usepackage{subfigure}
\usepackage{epstopdf}
\usepackage{enumerate}
\usepackage{tabularx}
\usepackage{amsthm}
\usepackage{xspace}
\usepackage{booktabs}
\usepackage{balance}
\usepackage{graphicx,pifont}
\usepackage{wrapfig}
\usepackage{multirow}
\usepackage{caption}
\usepackage[colorlinks=true,citecolor=red]{hyperref}

\ifCLASSINFOpdf
\else
\fi

\usepackage{amsmath}

\usepackage[cmintegrals]{newtxmath}

\definecolor{Red}{RGB}{0,0,0}
\definecolor{Yellow}{RGB}{243,140,10}

\newcommand\algorithmicprocedure{\textbf{procedure}}
\newcommand{\algorithmicendprocedure}{\algorithmicend\ \algorithmicprocedure}
\makeatletter
\newcommand\PROCEDURE[3][default]{%
  \ALC@it
  \algorithmicprocedure\ \textsc{#2}(#3)%
  \ALC@com{#1}%
  \begin{ALC@prc}%
}
\newcommand\ENDPROCEDURE{%
  \end{ALC@prc}%
  \ifthenelse{\boolean{ALC@noend}}{}{%
    \ALC@it\algorithmicendprocedure
  }%
}
\newenvironment{ALC@prc}{\begin{ALC@g}}{\end{ALC@g}}
\makeatother

\begin{document}

\title{Towards Complex Dynamic Physics System Simulation with Graph Neural ODEs}

\author{Guangsi Shi,
        Daokun Zhang,
        Ming Jin,\\
        Shirui Pan, \IEEEmembership{Senior Member,~IEEE}
        and Philip S. Yu,~\IEEEmembership{Life Fellow,~IEEE}

\IEEEcompsocitemizethanks{
\IEEEcompsocthanksitem Guangsi Shi is with the Department of Chemical and Biological Engineering, Faculty of Engineering, Monash University, Clayton, VIC 3800, Australia \protect\\
Email: guangsi.shi@monash.edu;
\IEEEcompsocthanksitem Daokun Zhang and Ming Jin are with the Department of Data Science and Artificial Intelligence, Faculty of IT, Monash University, Clayton, VIC 3800, Australia \protect\\
Emails: daokun.zhang@monash.edu and ming.jin@monash.edu;
\IEEEcompsocthanksitem Shirui Pan is with the School of Information and Communication Technology, Griffith University, South Port, QLD 4215, Australia \protect\\
Email: s.pan@griffith.edu.au;
\IEEEcompsocthanksitem Philip S. Yu is with Department of Computer Science, University of Illinois at Chicago, Chicago, IL 60607-7053, USA \protect\\
Email: psyu@uic.edu.
\IEEEcompsocthanksitem Corresponding authors: Daokun Zhang and Shirui Pan
} 
}
\markboth{Journal of \LaTeX\ Class Files,~Vol.~14, No.~8, August~2015}%
{Shell \MakeLowercase{\textit{et al.}}: Bare Demo of IEEEtran.cls for Computer Society Journals}

\IEEEtitleabstractindextext{%
\begin{abstract}
The great learning ability of deep learning facilitates us to comprehend the real physical world, making learning to simulate complicated particle systems a promising endeavour both in academia and industry. However, the complex laws of the physical world pose significant challenges to the learning based simulations, such as the varying spatial dependencies between interacting particles and varying temporal dependencies between particle system states in different time stamps, which dominate particles’ interacting behaviour and the physical systems’ evolution patterns. Existing learning based methods fail to fully account for the complexities, making them unable to yield satisfactory simulations. To better comprehend the complex physical laws, we propose a novel model – Graph Networks with Spatial-Temporal neural Ordinary Equations (GNSTODE) – that characterizes the varying spatial and temporal dependencies in particle systems using a united end-to-end framework. Through training with real-world particle-particle interaction observations, GNSTODE can simulate any possible particle systems with high precisions. We empirically evaluate GNSTODE’s simulation performance on two real-world particle systems, Gravity and Coulomb, with varying levels of spatial and temporal dependencies. The results show that GNSTODE yields significantly better simulations than state-of-the-art methods, showing that GNSTODE can serve as an effective tool for particle simulation in real-world applications.
\end{abstract}

}

\maketitle
\IEEEdisplaynontitleabstractindextext
\IEEEpeerreviewmaketitle

\section{Introduction}\label{sec:introduction}
\IEEEPARstart{S}{imulation} {has} become {an indispensable methodology} in many scientific and technological disciplines to comprehend the properties, behaviour, and dynamic changes of {real-world matters}. {Physical world simulations} are often {built on} the first principles~\cite{herfeld2021introduction}, 
{i.e., simulating according to the} established physics laws {without}  any {extra} assumptions.


\begin{figure}[t]
	\centering
	\includegraphics[width=9cm]{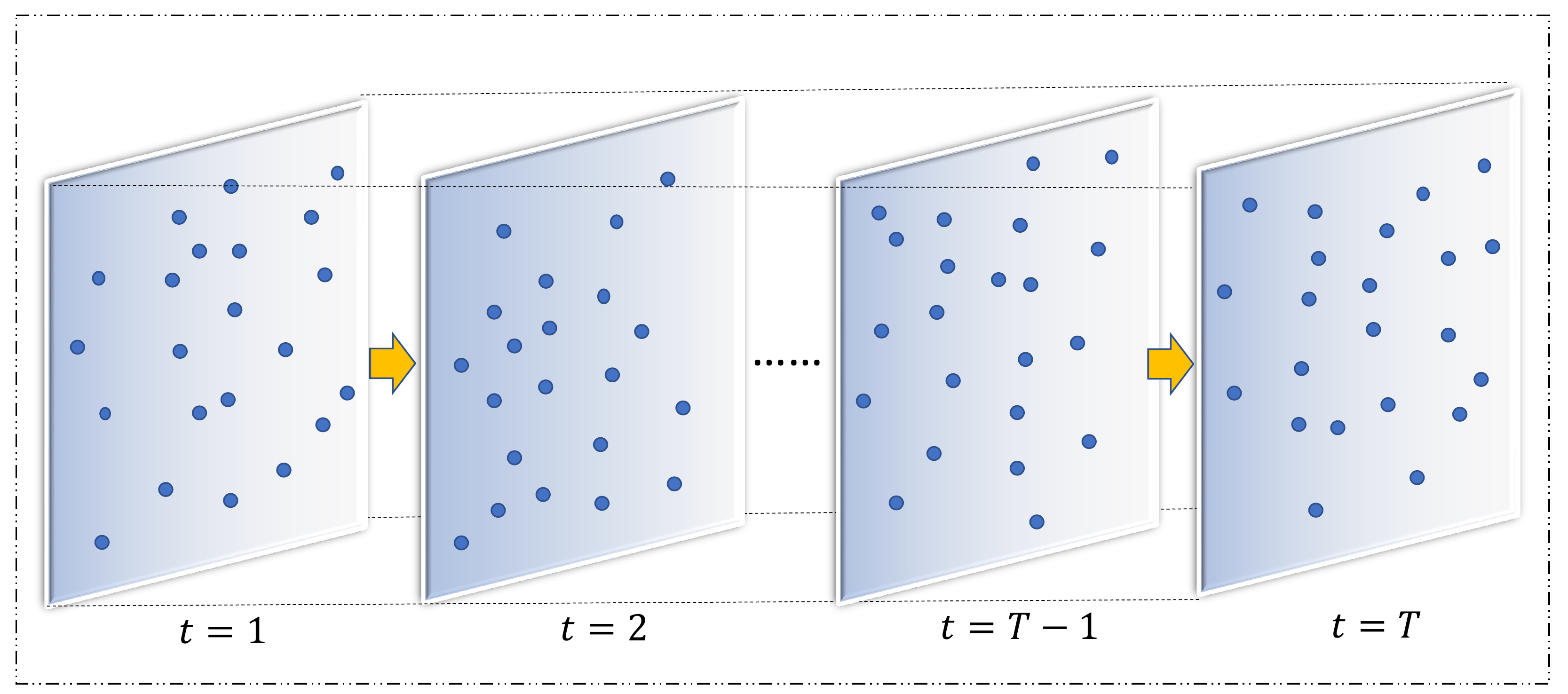}
	\vspace{-3mm}
 	\caption{The rollout of a particle system simulation.}
	\vspace{-3mm}
	\label{fig:rollout}
\end{figure}

Although the conventional Euler and Lagrangian simulation methods~\cite{subramaniam2013lagrangian}  have been employed by many disciplines, including pharmaceutical science and materials science, their simulation ability is still limited by some domain-specific challenges. First, the conventional physical simulation methods are usually developed on some prior expert knowledge describing the objective systems, like the
governing first principles, 
which makes them hard to generalize to the generic scenes with no prior knowledge.
Second, many physical systems are described by ordinary differential equations (ODEs) or partial differential equations (PDEs). As such, many trials are required to adjust the hyper-parameters of the ODE/PDEs to make them well fit real-world physical systems, which inevitably involves massive computational costs used to solve the complex ODE/PDEs.
Third, numerous simulators can only work for a single task and knowledge transfer across tasks is hard to achieve to yield high-quality simulations.
On the other hand, the learning-based simulation breaks away from the shackles of human knowledge and can expand the knowledge by uncovering more complex laws through learning from real-world observations. This makes the learning-based simulation a more promising solution to comprehending the complex physical world.


Some learning based algorithms have been developed
to achieve the complex physical simulations, including fluid simulation~\cite{brunton2020machine}, solid simulation~\cite{pfaff2020learning} and particle simulation~\cite{sanchez2020learning,Shlomi_2021}. Among others, graphs provide a straightforward data structure to describe the complex particle systems, with particles modelled as nodes and the interacting relations between particles described by edges. Based on the graph modeling of particle systems, Graph Neural Network (GNN)~\cite{9046288} has been leveraged as the simulation learner, due to its great expressive power. The GNN based simulation methods~\cite{li2019learning,sanchez2020learning,sanchez2018graph} leverage message passing mechanism to characterize the interactions between particles. Though competitive performance has been achieved in some cases, the applicability of GNN based simulation methods are still seriously challenged by the inherent complexities in the real-world particle systems:


\begin{itemize}

\item \textit{Challenge I: Varying Spatial Dependencies.}\label{challenge 1} In real-world physical systems, the random particle-particle interacting patterns result in the varying spatial dependencies between interacting particles. {Due to the irregular movements of particles, different particles can be surrounded by a varying number of neighbors, making particles have different centralities. With varying centralities, the interactions imposed by neighboring particles should be weighed differently to calculate the temporal dynamics of central particles at each time step.}
For particles with large centralities, interactions from close neighboring particles are enough to interpret their movement, while distant neighboring particles should also be considered for particles with small centralities. How to automatically determine the range of neighboring particles used to derive the driving forces of central particles is quite important to ensure accurate simulations. 


\vspace{0.1cm}    
\item \textit{Challenge II: Varying Temporal Dependencies.} \label{challenge 2} {Particle systems evolve with varying temporal patterns, i.e., the particle systems might change significantly in a show time slot or remain stable for a long time.
Existing learning based simulation methods mainly choose a fixed time step to roll out the simulation, by iteratively predicting the next time's state from the current time's state, as is shown in Fig. \ref{fig:rollout}. However, the rigid fixed-time-step prediction cannot capture the varying temporal dependencies of particle systems, without considering the possible dramatic changes occurring in the fixed time step. As a result, the one-step prediction inevitably involves many simulation errors,}
which will be further accumulated along the rollout of prediction and result in the model collapse dilemma~\cite{sanchez2020learning}. Therefore, the ability to adapt to the varying temporal dependencies of particle systems is critical to a learning based simulator for yielding stable and high-quality simulations.

\end{itemize}

{To effectively address the particle simulation chanllenges caused by varying spatial and temporal dependencies, we propose a novel learning based simulation model -- \underline{G}raph \underline{N}etworks with \underline{S}patial-\underline{T}emporal neural \underline{O}rdinary \underline{E}quations (GNSTODE) -- that leverages neural ordinary equations to learn varying spatial correlations between particles with varying distances and to model the complex system evolving dynamics. Compared with the existing methods~\cite{sanchez2020learning,martinkus2021scalable,sanchezgonzalez2019hamiltonian} that use the rigid increment-adding operators (e.g., the Runge-Kutta integrator~\cite{sanchezgonzalez2019hamiltonian}) to predict the next time stamps' system states, the GNSTODE is endowed with the ability to adapt to the varying spatial and temporal dependencies by making predictions through performing the integral operation with regard to the learned spatial and temporal derivatives.}

{On the spatial domain, a common GNN based strategy~\cite{9046288} to calculate the temporal dynamics at each time stamp is to use the iterative graph convolution operation to increasingly account for the effects of distant neighboring particles. To better adapt to varying spatial dependencies, we infer the temporal dynamics at each time step by numerically solving a neural Ordinary Differential Equation (ODE) on the spatial domain, where we use a GNN to model the derivative of the temporal dynamics with regard to the neighborhood radius. As the solution to the neural ODE, the final temporal dynamics are obtained by integrating the GNN based derivative along the neighborhood radius.}
{Through learning from real-world data, the modelled derivative can automatically explore the varying importance of neighboring particles with varying distances to the central particles, by adaptively assigning large derivative values to small neighborhood radiuses while making the large neighborhood radiuses have small derivative values.}

{On the temporal domain, to make the model adaptive to the varying temporal dependencies within a fixed time step, we use a neural network to parameterize a continuous temporal dynamics function, that interpolates the known temporal dynamics modeled by spatial neural ODE at finite time points. Based on the fact that the modeled continuous temporal dynamics function is the derivative of the system state with regard to time, we predict the next time's system state by solving another neural ODE on the temporal domain, by integrating the continuous temporal dynamics from the last time stamp to the current time stamp. Through using continuous temporal dynamics to model any possible state change trends within a fixed time step, the proposed GNSTODE can make an accurate prediction for the next time's state. The accurate time point-wise prediction makes no simulation errors accumulated along the simulation rollout, which guarantees the stability and accuracy of the overall simulation.}


{In the GNSTODE model, to make the spatial and temporal ODEs well collaborate with each other, we integrate them into a united end-to-end learning framework. We also develop an algorithm for training the proposed GNSTODE model efficiently, in which numerical ODE solvers are leveraged to achieve the fast feedforward prediction, while the adjoint sensitivity method~\cite{pontryagin1987mathematical} is employed for instant error backpropagation. We conduct extensive experiments on simulating the particle systems with varying spatial and temporal dependencies. The results show that the proposed GNSTODE model achieves significantly better simulation performance than state-of-the-art learning based simulation methods in all cases. In summary, the contribution of this paper is threefold:}

\begin{itemize}
    
\item{To the best of our knowledge, we are the first to analyze the two key challenges for learning to simulate particle systems: the varying spatial and temporal dependencies. This not only well motivates the proposed approach but also provides a high-level instruction for the follow-up development of the learning based simulation models.}

\item{{We propose a novel learning based simulation model GNSTODE that is able to adapt to varying spatial and temporal dependencies in particle systems, by creatively leveraging two coupled ODEs to model the varying dependencies in spatial and temporal domains. An efficient training algorithm is also designed to make sure GNSTODE is practical to real-world simulations.}}
\item{To verify the effectiveness of the proposed GNSTODE model, we conduct extensive simulation experiments on real-world particle systems with different spatial and temporal dependencies. The experimental results show that the proposed GNSTODE model consistently outperforms state-of-the-art learning based simulation models in all cases by large margins.}

\end{itemize}

The rest of this paper is organized as follows. In Section \ref{sec:related work}, we first review the related work. Then, 
{the problem of learning to simulate is formally defined in}
in Section \ref{sec:problem formulation}. 
{Section \ref{sec:preliminaries} then reviews the required preliminaries}. {After that, Section \ref{sec:methodology} details the proposed GNSTODE model. Extensive experiments are then presented in Section \ref{sec:experiments}. Finally, we conclude this paper in Section \ref{sec:conclusion}. }

\section{Related Work}\label{sec:related work}
{In this section, we review three streams of related work: physics system simulation, Graph Neural Networks (GNNs) and Neural Ordinary Equations (Neural ODEs).}
\subsection{Physics System Simulation}\label{subsec:traditional simulation}

Traditional physics system simulation is achieved through modeling the system with basic physics laws. It mainly uses two rationales: 1) the Euler rationale that treats the physics systems as continuums, with representative models of Computational Fluid Dynamics (CFD)~\cite{ferziger2002computational} and Finite Element Method (FEM)~\cite{zienkiewicz2005finite}; 2) the Lagrangian rationale that models the physics systems as discrete matters (particles), with examples of Discrete Element Method (DEM)~\cite{zhu2007discrete}, Lattice Boltzmann Method (LBM)~\cite{chen1998lattice} and Smooth Particle Hydrodynamics (SPH)~\cite{monaghan1992smoothed}, etc. As many real-world systems are described by particles, in this paper, we consider the simulation under the Lagrangian rationale, i.e., particle system simulation. 

As the underlying physics laws are hard to fully be uncovered, traditional particle system simulation methods fail to yield satisfactory simulations for many real-world systems. As a more advanced simulation paradigm, learning based particle system simulation has been proposed to perceive the underlying physics laws from real-world data with machine learning. 
Learning based particle system simulation mainly leverages neural networks to model the interactions between particles, with the purpose of capturing the relational inductive biases~\cite{battaglia2018relational}. Graph Interaction Networks (GINs)~\cite{battaglia2016interaction} is the pioneer of learning to simulate particle systems. Neural Relational Inference (NRI)~\cite{kipf2018neural} is then proposed to consider different interacting types. HOGN~\cite{sanchezgonzalez2019hamiltonian} imports the Hamiltonian mechanics~\cite{greydanus2019hamiltonian} as physics informed inductive biases into INs for more accurate particle system simulation. To better characterize the complex particle systems, DPI-Nets~\cite{li2019learning} try to capture the hierarchical and long-range interactions between particles with the proposed Propagation Networks. GNS~\cite{sanchez2020learning} models the hierarchical and long-range particle-particle interactions by the iterative massage-passing mechanism~\cite{gilmer2017neural}. GNS variants are then developed by improving its scalability~\cite{martinkus2021scalable} and adapting it to the mesh-based simulation~\cite{pfaff2020learning}. To make neural models reason like humans, VGPL~\cite{li2020visual} reinforces the learning based particle system simulation with visual and dynamics priors. 

However, the existing learning based particle system simulation does not account for the varying spatial and temporal dependencies in real-world particle systems, limiting their performance. In this paper, we propose to use Graph Neural ODEs to characterize the varying spatial and temporal dependencies, proving a more effective method for real-world particle system simulation.

\subsection{Graph Neural Networks (GNNs)}\label{subsec:GNN}

GNNs are one of the most powerful neural network architectures for learning from graph structured data~\cite{9046288,zhang2018network}, which are first proposed by \cite{gori2005new}. 
After that, a number of spectral GNNs~\cite{ defferrard2016convolutional} are then proposed, by performing feature filtering on graph spectral domains. Many spatial variants have also been developed for GNNs, like Graph Convolution Network (GCN)~\cite{gcn_kipf2017semi} that uses neighborhood aggregation to update node representations, and Graph Attention Network (GAT)~\cite{gat_ve2018graph} that considers the attention weights between neighboring nodes when performing neighborhood aggregation. Among others, GNS~\cite{sanchez2020learning} provides a general GNNs based framework for particle system simulation, by modeling particles as nodes and characterizing the interactions between particles by edges. Recent research on GNNs includes improving GNNs' scalability for handling large graphs~\cite{wu2019simplifying,hamilton2017inductive}, developing more powerful neighborhood aggregation operators 
\cite{xu2018powerful,klicpera2019diffusion} and learning informative graph representations with self-supervised learning~\cite{pan2019learning,liu2021graph}, etc. GNNs have been applied to solve many real-world problems in various disciplines, including time-series prediction~\cite{wu2019graph, wu2020connecting}, anomaly detection~\cite{liu2021anomaly}, hyperspectral image classification~\cite{wan2020hyperspectral}, knowledge graph completion~\cite{9416312}, molecular and material property prediction~\cite{gilmer2017neural,karamad2020orbital} and protein structure prediction~\cite{jumper2021highly}, etc.

In this paper, we propose a novel framework to make the best of GNNs' expressive power for accurate particle system simulation, by making the GNNs based particle-particle interaction modeling well interplay with Neural ODEs to capture the complex temporal and spatial dependencies in real-world particle systems.


\subsection{Neural ODEs}\label{subsec:neural odes}

Neural ODEs~\cite{chen2018neural} are a new class of deep neural networks, which model deep neural network transformations as the numerical solutions of ODEs. They have been applied to construct continuous-depth residual networks~\cite{chen2018neural}, with much lower memory cost than deep residual networks~\cite{he2016deep}. Neural ODEs also provide an elegant way to model the continuous-time latent states for time series~\cite{chen2018neural}. As an example, ODE-RNNs \cite{rubanova2019latent} augment Recurrent Neural Networks (RNNs) with Neural ODEs to accurate model hidden states for irregularly-sampled time series. In addition to time series, Neural ODEs have also been applied to model other temporal systems. LG-ODE~\cite{huang2020learning} uses Neural ODEs to model the multi-agent dynamic systems with known graph structures and irregularly-sampled partial observations.
TrajODE~\cite{liang2021modeling} applies Neural ODEs to model the continuous-time trajectory dynamics in the noisy spatial trajectory data.
Besides, Neural ODEs have also been leveraged to model spatio-temporal point processes~\cite{chen2020neural}, learn event functions~\cite{chen2020learning}, and dynamic graph\cite{jin2022multivariate, jin2022neural}. 

The success of Neural ODEs on modeling temporal dynamic systems inspires us to leverage them to capture the complex continuous-time temporal dynamics in the real-world particle systems. To fully unleash the expressive power of Neural ODEs, we also apply Neural ODEs to characterize the dynamics in the spatial domain, i.e.,  the varying spatial dependencies between particles with varying distances. 

\section{Problem Formulation}\label{sec:problem formulation}

For the problem of learning to simulate, we are given a collection of observed particle state trajectories $\{\mathcal{X}\in\mathbb{R}^{n\times d\times T}\}$ as training samples, with $\mathcal{X}=\{X_1,X_2,\cdots,X_T\}$, where $n$ is the number of particles in the particle system, $d$ is the number of particle features (e.g., coordinates, velocities, accelerations, masses and electric charges) for describing particles' states, $T$ is the number of time stamps of the trajectories, and $X_t\in\mathbb{R}^{n\times d}$ is the particle system state at time stamp $t$, with its $i$-th column $\bm{x}_{t,i}\in\mathbb{R}^d$ representing the feature vector of the $i$-th particle. Using training samples, we aim to learn a simulator, $\mathcal{S}: \mathbb{R}^{n\times d}\rightarrow \mathbb{R}^{n\times d}$, which is able to predict the particle system state at time stamp $t$ with its previous state at time stamp $t-1$, i.e., $\hat{X}_{t+1}=\mathcal{S}(X_t)$. The training objective is to minimize the difference between the predicted state $\hat{X}_{t+1}$ and the ground-truth state $X_{t+1}$. 

To construct the simulator $\mathcal{S}$, we first need to learn a temporal dynamics function $\mathcal{D}: \mathbb{R}^{n\times d}\rightarrow \mathbb{R}^{n\times d}$, for predicting the temporal dynamics at each time stamp: $D_t=\mathcal{D}(X_t)$. The temporal dynamics $D_t\in\mathbb{R}^{n\times d}$ describes the trend by which the particle system state $X_t$ evolves to the next time stamp's state $X_{t+1}$. The simulator $\mathcal{S}$ is then obtained by updating the particle system state $X_t$ with its temporal dynamics $D_t$: $\mathcal{S}(X_t)=\mathcal{F}(X_t,D_t)=\hat{X}_{t+1}$, where $\mathcal{F}: \mathbb{R}^{n\times d}\times\mathbb{R}^{n\times d}\rightarrow\mathbb{R}^{n\times d}$ is the updating function. 

At each time stamp $t$, we design a spatial Neural ODE to formulate the temporal dynamics function $\mathcal{D}$, by considering the varying spatial dependencies between particles. For characterizing the spatial dependencies between particles, a spatial graph is constructed according to the particle coordinates, $G_t=(V_t, E_t, X_t)$, where $V_t$ is the set of nodes with each node corresponding to a particle, $E_t\subseteq V_t\times V_t$ is the set of edges with each edge connecting two particles having a distance smaller than a predefined threshold, and $X_t\in\mathbb{R}^{n\times d}$ describes the particle features. In the spatial Neural ODE based temporal dynamics function formulation, the Graph Interaction Network (GIN)~\cite{battaglia2016interaction} will be used for effectively capturing the interactions between particles. 

To capture the varying temporal dependencies between consecutive time stamps, we design a temporal Neural ODE to formulate the updating function $\mathcal{F}$. Instead of using a simple one-step updating operation (e.g., addition), the temporal Neural ODE extrapolates the continuous temporal dynamics between consecutive time stamps, so that the accumulative updating along a continuous time slot can be achieved to accurately predict the next time stamp's state. 

With the well-trained simulator $\mathcal{S}$, given the initial state of a particle system $X_1$, we can simulate its state evolving trajectory $\hat{\mathcal{X}}=\{X_1,\hat{X}_2,\cdots,\hat{X}_T\}$, by sequentially applying the simulator $\mathcal{S}$ at each time stamp, with $\hat{X}_2=\mathcal{S}(X_1)$ and $\hat{X}_{t+1}=\mathcal{S}(\hat{X}_t)$ for $t\geq 2$. As the simulator $\mathcal{S}$ is designed for effectively capturing the particle systems' complex evolving patterns, i.e., the varying spatial and temporal dependencies, we can expect that the simulated particle state trajectory $\hat{\mathcal{X}}$ will be consistent with the ground-truth state trajectory $\mathcal{X}$.

\section{Preliminaries}\label{sec:preliminaries}
\begin{figure*}[tbp]
	\centering
	\includegraphics[width=1.0\textwidth]{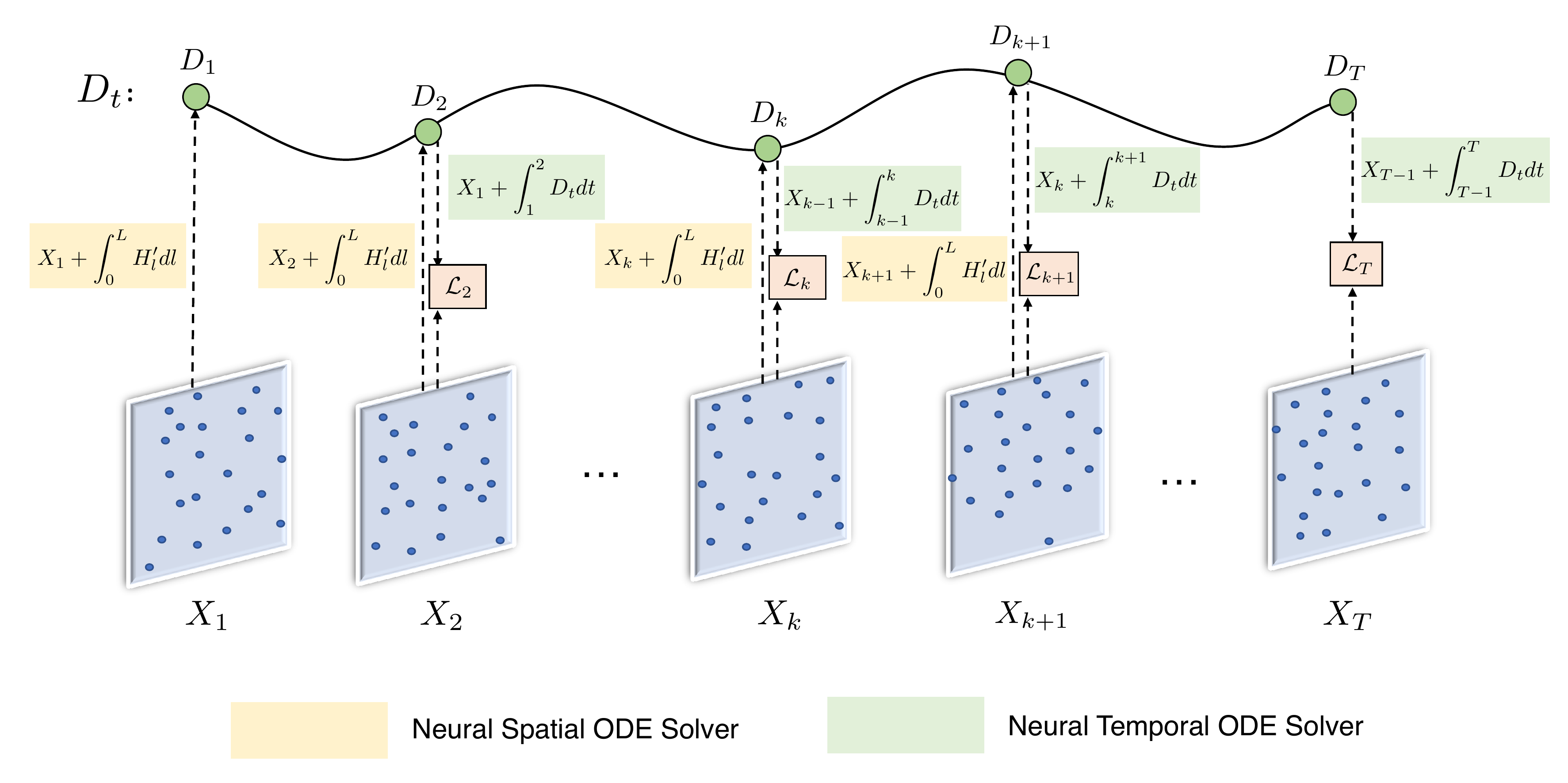}
	\caption{The overall framework of GNSTODE. The framework is composed of two key components: the neural spatial ODE component is used to predict current temporal dynamics, while the neural temporal ODE component is employed to predict the next time stamp's system state.}
	\label{fig:framework}
\end{figure*}

In this section, we review the preliminaries on Graph Interaction Network (GIN)~\cite{battaglia2018relational} and Neural ODEs~\cite{chen2018neural}. 
\subsection{Graph Interaction Network (GIN)} 

At each time stamp $t$, by formulating the particle-particle interaction relations with a particle interaction network $G_t=(V_t, E_t, X_t)$, we can use the Graph Interaction Network (GIN)~\cite{battaglia2018relational} to learn informative particle representations that well encode the particle-particle interaction patterns. The particle temporal dynamics can then be predicted from particle representations by performing a non-linear transformation. GIN is a stacked multi-layer neural network, with each layer performing a message-passing operation. By denoting the $l$-th-layer input representation of the $i$-th particle as $\bm{h}_{l,i}\in\mathbb{R}^{d_l}$, in the $l$-th layer, the representation is updated as 
\begin{equation}\label{message-passing}
\begin{aligned}
&\bm{m}_{l,i} = \sum_{j\in\mathcal{N}(i)}\mathcal{M}_{l}(\bm{h}_{l,i},\bm{h}_{l,j},\bm{e}_{i,j}),\\
&\bm{h}_{l+1,i} = \mathcal{U}_{l}(\bm{h}_{l,i},\bm{m}_{l,i}),
\end{aligned}
\end{equation}where $\mathcal{N}(i)$ is the set of particle $i$'s neighboring particles; $\bm{e}_{i,j}\in\mathbb{R}^{d_e}$ represents the features of the edge connecting particles $i$ and $j$, which can be the distance or interacting forces between them; $\mathcal{M}_{l}(\cdot,\cdot,\cdot)$ is the $l$-th-layer function for calculating the messages between interacting particles; $\bm{m}_{l,i}$ represents the aggregated messages passed to particle $i$ in the $l$-th layer; and $\mathcal{U}_{l}(\cdot,\cdot)$ is the function for updating particle representations with aggregated messages in the $l$-th layer. 

With the message-passing mechanism serving as an effective relational inductive bias, GIN can accurately predict particle temporal dynamics from the particle-particle interaction patterns. For mathematical convenience, we denote input particle representations of the $l$-th layer as $H_l\in\mathbb{R}^{n\times d_l}$ with its $i$-th column being the $i$-th particle's representation $\bm{h}_{l,i}$, and set $H_0=X_t$, then we reformulate the message passing operation in Eq. (\ref{message-passing}) as 
\begin{equation}
H_{l+1}=\mathrm{GIN}_{l}(H_l).    
\end{equation}

\subsection{Neural ODEs}  

Many deep neural networks (e.g., residual networks and recurrent neural networks) work through iteratively operating a transformation on the networks' hidden states: 
\begin{equation}
\bm{z}_{t+1} = \bm{z}_{t} + f_{t}(\bm{z}_t),
\end{equation}where $t\in\{0,1,\cdots,T\}$ is the layer number, $\bm{z}_t\in\mathbb{R}^d$ is the hidden state at the $t$-th layer and $f_t(\cdot)$ is the $t$-th layer's transformation function.

Neural ODEs~\cite{chen2018neural} reformulate the discrete updating as a continuous integral operation, by formulating the relation between hidden state $\bm{z}_t$ and layer number $t$ with an ODE:
\begin{equation}
\frac{d \bm{z}_t}{dt}=f(\bm{z}_t,t), 
\end{equation} where $f(\cdot,\cdot)$ is the layer-dependent transformation function parameterized by a neural network. Given the input state $\bm{z}_0$, the hidden state at the $t$-th layer $\bm{z}_t$ is calculated by 
\begin{equation}
\bm{z}_t=\bm{z}_{0}+\int_{0}^{t}f(\bm{z}_{\tau},\tau)d\tau,
\end{equation}which can be solved by numerical ODE solvers, such as Euler and Runge-Kutta~\cite{chen2018neural}. Similar to traditional deep neural networks, the learnable parameters in $f(\cdot,\cdot)$ are updated by error backpropagation, where their gradients can be fast computed by the adjoint sensitivity method~\cite{pontryagin1987mathematical}.

\section{Methodology}\label{sec:methodology}


In this section, we detail the proposed GNSTODE framework to learn the simulator function $\mathcal{S}$ for predicting the next time stamp's system state according to the current state. To yield high-quality simulation, the proposed GNSTODE framework effectively models the varying spatial dependencies via a neural spatial ODE component and captures the varying temporal dependencies with a neural temporal ODE component.

\subsection{Overall Framework} \label{subsec:model framework}  


Fig. \ref{fig:framework} shows the overall GNSTODE learning framework. At each time stampt $t$, a neural spatial ODE is used to construct the temporal dynamics function $\mathcal{D}:\mathbb{R}^{n\times d}\rightarrow\mathbb{R}^{n\times d}$, for inferring the current temporal dynamics $D_t$ from the current system state $X_t$. The neural spatial ODE is solved through an integral along the varying distances between neighboring particles and central particles, to characterize the varying interacting effects between particles with varying distances. On the other hand, a neural temporal ODE is used to construct the updating function $\mathcal{F}:\mathbb{R}^{n\times d}\times\mathbb{R}^{n\times d}\rightarrow\mathbb{R}^{n\times d}$, for predicting the current system state $\hat{X}_t$ from the last time stamp's state $X_{t-1}$ and the last time stamp's temporal dynamics $D_{t-1}$. Through extrapolating the continuous temporal dynamics function between the time stamps $t-1$ and $t$, the temporal neural ODE can effectively capture the complex accumulative updating effects between consecutive time stamps. The reconstruction loss at the time stamp $\mathcal{L}_t$ is then calculated through comparing the predicted system state $\hat{X}_t$ with the ground truth state $X_{t}$. The model is trained by minimizing the overall construction loss $\mathcal{L}$, which is obtained by summing up the reconstruction losses at all time stamps. 


\subsection{Neural Spatial ODE Component}\label{subsec:NSODE}

At each time stamp $t$, by representing the particle system state $X_t$ as a spatial graph $G_t=(V_t,E_t,X_t)$, we use the Graph Interaction Network (GIN)~\cite{battaglia2018relational} to model the interactions between particles. We aim to make the best of GIN's expressive power to predict the temporal dynamics $D_t$ accurately. GIN embeds the complex interactions between neighboring particles into particle latent representations with the iterative message-passing operation:
\begin{equation}\label{GIN_layer}
H_{l+1}=\mathrm{GIN}_{l}(H_l), 
\end{equation}where $\mathrm{GIN}_{l}(\cdot)$ is the $l$-th GIN layer's message passing operation defined by Eq. (\ref{message-passing}) and $H_l$ is the input particle representations of the $l$-th GIN layer with $H_0=X_t$. 

To avoid over-smoothing and make the hidden representations more indicative to particle original features, by augmenting a skip-connection operation, we reformulate the GIN's layer-wise updating in Eq. (\ref{GIN_layer}) as
\begin{equation}\label{GIN_skip_layer}
H_{l+1}=H_l+\mathrm{GIN}_{l}(H_l).
\end{equation}By regarding the operation in Eq. (\ref{GIN_skip_layer}) as the discrete approximation of a continuous-layer updating, we can write the derivative of the hidden representations $H_l$ with regard to the layer number $l$, $H^{\prime}_l$, as,
\begin{equation}\label{spatial_ode_formulation}
H^{\prime}_l=\frac{d H_l}{dl}=\mathrm{GIN}_l(H_l), 
\end{equation}where the GIN layer-wise operation $\mathrm{GIN}_l(\cdot)$ is used to formulate the hidden representation derivative. Assuming the upper bound of layer number is set to $L$, the final-layer particle representations can be obtained by solving the ODE problem in Eq. (\ref{spatial_ode_formulation}) as:
\begin{equation}\label{spatial_ode_solution}
H_L=H_0+\int_{0}^{L}H^{\prime}_{l}dl=X_t+\int_{0}^{L}H^{\prime}_{l}dl.
\end{equation}In implementation, we set $L=1$, which warrants $L$ to be automatically re-scaled to arbitrary values with the learnable parameters in $H^{\prime}_l$.

The temporal dynamics at time stamp $t$, $D_t$, is finally obtained by performing a multi-layer neural network based transformation $\mathrm{NN}(\cdot)$ on $H_L$:
\begin{equation}\label{temporal_dynamics}
D_t=\mathrm{NN}(H_L).
\end{equation}

\subsection{Neural Temporal ODE Component}\label{subsec:NTODE}

At each time stamp $t$, given the system state $X_t$ and the inferred temporal dynamics $D_t$, we can predict the next time stamp's state with the updating function $\mathcal{F}$:
\begin{equation}
\hat{X}_{t+1} = \mathcal{F}(X_t,D_t). 
\end{equation}A common implementation of $\mathcal{F}$ is to use the simple addition operator: 
\begin{equation}\label{naive_temporal_updating}
\hat{X}_{t+1}=X_t+D_t.
\end{equation}However, such a naive updating operation cannot account for the complex accumulative effects between time stamps $t$ and $t+1$. To achieve continuous-time updating, we model the relation between the particle system state and time as an ODE:
\begin{equation}\label{temporal_ODE_formulation}
\left\{
\begin{aligned}
&\frac{d\hat{X}_\tau}{d\tau}=D_\tau, \;\;\tau\in[t,t+1],\\
&D_{\tau}=D_t\;\;\mathrm{for}\;\tau=t.
\end{aligned}
\right.
\end{equation}Here, we use a neural network to parameterize the temporal dynamics function $D_{\tau}$ with input time variable $\tau\in[t,t+1]$. The next time stamp's state is then predicted by solving Eq. (\ref{temporal_ODE_formulation}) as
\begin{equation}\label{temporal_ode_solution}
\hat{X}_{t+1}=X_t+\int_{t}^{t+1}D_{\tau}d\tau.
\end{equation}

\subsection{Loss Function}
For each time stamp $t\geq 2$, by comparing the predicted particle system state $\hat{X}_t$ with the ground-truth state $X_t$, we calculate the reconstruction loss as
\begin{equation}
\mathcal{L}_t = \Vert\hat{X}_t-X_t\Vert_{\mathrm{F}}^2,
\end{equation}where $\Vert\cdot\Vert_\mathrm{F}$ is the Frobenius norm. The overall reconstruction loss of the whole state trajectory is obtained by summing up the reconstruction losses at all time stamps:
\begin{equation}\label{overall_loss}
\mathcal{L}=\sum_{t=2}^{T}\mathcal{L}_t.
\end{equation}The parameters of the proposed GNSTODE is optimized by minimizing the reconstruction loss $\mathcal{L}$ of each training state trajectory. We use error propagation to update model parameters, where the Runge-Kutta method~\cite{chen2018neural} is used to numerically calculate the ODE solutions in Eq. (\ref{spatial_ode_solution}) and (\ref{temporal_ode_solution}) in the feedforward process, and the gradients of the ODE parameters are fast calculated by the adjoint sensitivity method~\cite{pontryagin1987mathematical} in the backpropagation process.

\noindent\textbf{Algorithm Description and Time Complexity}. Algorithm \ref{algo: overall algorithm} describes the overall workflow for training the proposed GNSTODE model. We first initialize model parameters with random numbers. The model parameters are then updated by stochastic gradient descent: randomly sample a batch of training samples at each iteration, and update model parameters with gradient descent by reducing the corresponding reconstruction loss.
Taking the maximum number of epochs \textit{max\_epoch} as a constant, the time complexity of Algorithm \ref{algo: overall algorithm} is $O((nd^2+nds_1+nds_2)T|\mathcal{X}|)$, where $s_1$ and $s_2$ are respectively the numbers of steps adopted by the Runge-Kutta method for calculating the ODE solutions in Eq. (\ref{spatial_ode_solution}) and (\ref{temporal_ode_solution}), and $|\mathcal{X}|$ is the number of training samples in the training set $\mathcal{X}$. 

\begin{algorithm}[t]
	\caption{The algorithm for training GNSTODE}
	\label{algo: overall algorithm}
    \textbf{Input}: The particle state trajectory training set $\{\mathcal{X}\in\mathbb{R}^{n\times d\times T}\}$.\\
    \textbf{Output}: The trained GNSTGODE model. \\ \vspace{-4mm}
    \begin{algorithmic}[1]
        \STATE Initialize model parameters with random numbers.
		\FOR{\textit{epoch} in 1, 2, $\cdots$, \textit{max\_epoch}}
			\STATE $\mathcal{B} \leftarrow$ Randomly split $\mathcal{X}$ into batches of size $B$;
            \FOR{each batch in $\mathcal{B}$}
            \STATE Calculate temporal dynamics at time stamps $t=2,3,\cdots,T$ with Eq. (\ref{temporal_dynamics});\
            \STATE Predict particle system states at time stamps $t=2,3,\cdots,T$ with Eq. (\ref{temporal_ode_solution});\
            \STATE Update model parameters by reducing the reconstruction loss in Eq. (\ref{overall_loss}) with gradient descent;\
            \ENDFOR
		\ENDFOR
		\STATE \textbf{return} the trained GNSTGODE model.
	\end{algorithmic}
\end{algorithm}

\section{Experiments}\label{sec:experiments}
In this section, we conduct experiments to verify the effectiveness of the proposed GNSTODE model for particle system simulation. We first introduce the datasets used for experiments and the experimental setup. Then, we demonstrate the experimental results, including the performance comparison with baselines, ablation study and parameter sensitivity study.

\renewcommand{\arraystretch}{1.15}
\begin{table}
	\centering
    \caption{Dataset statistics.}
	\resizebox{0.9\columnwidth}{!}{
	\begin{tabular}{@{}c|c|c@{}}
		\toprule
        Dataset & \#Particles & \#Particle Features \\
        \midrule
		\multirow{3}{*}{Gravity}  & 20    & 5   \\
                                  & 100   & 5\\
                                  & 500   & 5\\
        \midrule
		\multirow{3}{*}{Coulomb}  & 20     & 6  \\
                                  & 100    & 6  \\
                                  & 500    & 6  \\
		\bottomrule
	\end{tabular}
	}
	\label{table:dataset}
\end{table}

\subsection{Benchmark Datasets} \label{subsec:exp_dataset}

We evaluate the proposed GNSTODE model on six widely used particle simulation benchmark datasets~\cite{martinkus2021scalable}, including two types of particle systems (Gravity and Coulomb) and three different particle scales (20, 100, 500) for each type.

In the Gravity particle system, each particle $i$ is descried by a five-dimensional feature vector $[m_i, x_i, y_i, \dot{x}_i, \dot{y}_i]$, where $m_i$ is the mass of the particle and $\bm{x}_i=[x_i,y_i]$ is the coordinate of the particle in the two-dimensional Euclidean space and $\dot{\bm{x}}_i=[\dot{x}_i,\dot{y}_i]$ is the velocity of the particle. Differently, each particle $i$ is described by a 6-dimensional feature vector $[m_i, c_i, x_i, y_i, \dot{x}_i, \dot{y}_i]$ in the Coulomb system, with an additional feature dimension -- the electric charge $c_i$. The statistics of the particle systems is summarized in Table \ref{table:dataset}.

For each particle system with varying types and varying scales, we respectively simulate 100 training, 20 validation and 20 testing trajectories with 200 time stamps for each, using the Leapfrog simulator~\cite{martinkus2021scalable}. The Leapfrog simulator leverages Newton's second law to update particle coordinates and velocities, and respectively utilize the Gravity and Coulomb forces to formulate particle accelerations in the Gravity and Coulomb systems. In the Gravity system, the particle acceleration is calculated as
\begin{equation}
    \bm{a}_i = - G \sum_{j \neq i} \frac{m_j}{\|\bm{x}_i-\bm{x}_j\|_2^2 }, 
    \label{equ:gravity a}
\end{equation}where $G$ is the gravity constant. In the Coulomb system, the acceleration is instead computed as
\begin{equation}
    \bm{a}_i = k \frac{1}{m_i} \sum_{j \neq i} \frac{c_i \cdot c_j}{\|\bm{x}_i-\bm{x}_j\|_2^2}, 
    \label{equ:coulomb c}
\end{equation}where $k$ is the Coulomb constant. 


For each simulation, particles' initial coordinates are determined by the random sampling from a two-dimensional uniform random distribution with intensity 0.42 (the average particle number at per $1\times 1$ square);
particle masses are uniformly set to 1; particle velocities are initialised by sampling over the uniform distribution on $(-1,1)$; particle charges are initialised by the random sampling from the uniform distribution on $(0.5,1.5)$ for the Coulomb system; the Gravity and Coulomb constants are respectively set as $G=2$ and $k=2$; and the time step is set as $0.01$. 

\subsection{Experimental Settings} \label{subsec:exp_setting}  

To provide a comprehensive evaluation on the simulation performance of the proposed GNSTODE model, we configure three different experimental settings:
\begin{itemize}
    \item \textbf{Varying Scales}. This setting evaluates the robustness of the GNSTODE model against varying particle scales.  As is described, the three different particle scales (20, 100, 500) are respectively used for the Gravity and Coulomb systems for the evaluation purpose. 
    \item \textbf{Varying Intensities}. This setting aims to mimic the varying spatial dependencies between particles. Fixing the particle system scale to 20, we respectively simulate three different system variants with initial particle intensities 0.083 0.42 and 1.63 for the Gravity and Coulomb systems. 
    \item \textbf{Varying Time Steps}. This setting mimics the varying temporal dependencies. On the Gravity and Coulomb particle trajectories with scale 20 and initial particle intensity 0.42, we respectively perform three downsampling operations by keeping only every 5th, 10th and 20th time stamps and benchmark on the downsampled trajectories. 
\end{itemize}



\begin{table*}[t]
\scriptsize
\centering
\caption{Simulation performance comparison on the Gravity and Coulomb systems under the setting of varying scales. The best and second best performers are respectively highlighted by the \textbf{boldface} and \underline{underline}.
}
\resizebox{1.8\columnwidth}{!}{
\begin{tabular}{l|l|c|c c c c c}
\toprule
Dataset & Metrics $\downarrow$ & \multicolumn{1}{l|}{Scale} & \multicolumn{1}{l}{GNS} & \multicolumn{1}{l}{HOGN} & \multicolumn{1}{l}{HDGN} & \multicolumn{1}{l}{HHOGN} & \multicolumn{1}{l}{GNSTODE} \\ 
\midrule
\multirow{6}{*}{Gravity} & \multirow{3}{*}{RMSE} & 20 & 0.8654 & 1.1053 & \underline{0.8114} & 1.1211 & \textbf{0.7053}\\  
 & & 100 & 1.7098 & 1.7120 & \underline{1.0471} & 1.5190 & \textbf{0.9937} \\  
 & & 500 & 1.6813 & 1.7943 & \underline{1.0033} & 1.5533 & \textbf{0.7447} \\ \cline{2-8} 
 & \multirow{3}{*}{\begin{tabular}[c]{@{}l@{}}Energy\\Error\end{tabular}} & 20 & 0.0659 & 0.4227 & \underline{0.0584} & 0.4237 & \textbf{0.0167} \\ 
 & & 100 & 0.4116 & 0.2160 & \underline{0.0466} & 0.1918 & \textbf{0.0199} \\  
 & & 500 & 0.1268 & 0.2201 & \underline{0.0134} & 0.0858 & \textbf{0.0029} \\ 
 \midrule
\multirow{6}{*}{Coulomb} & \multirow{3}{*}{RMSE} & 20 & 0.8471 & 1.1022 & 0.7770 & \underline{0.7475} & \textbf{0.6797} \\ 
 & & 100 & 0.6519 & 0.7459 & \underline{0.4968} & 0.7857 & \textbf{0.4505} \\  
 & & 500 & 0.6921 & 0.7857 & 0.5562 & \underline{0.5442} & \textbf{0.3334} \\ \cline{2-8} 
 & \multirow{3}{*}{\begin{tabular}[c]{@{}l@{}}Energy\\Error\end{tabular}} & 20 & 0.0642 & 0.4071 & \underline{0.0248} & 0.4301 & \textbf{0.0116} \\ 
 & & 100 & 0.7375 & \underline{0.3810} & 0.8405 & 0.4075 & \textbf{0.3273} \\ 
 & & 500 & 0.7373 & 0.8901 & \underline{0.2175} & 0.8614 & \textbf{0.0423} \\ 
 \bottomrule
\end{tabular}}
\label{table: scale}
\end{table*}

\begin{table*}[t]
\scriptsize
\centering
\caption{Simulation performance comparison on the Gravity and Coulomb systems under the setting of varying intensities. The best and second best performers are respectively highlighted by the \textbf{boldface} and \underline{underline}.
}
\resizebox{1.8\columnwidth}{!}{
\begin{tabular}{l|l|c|c c c c c}
\toprule
Dataset & Metrics $\downarrow$ & Intensity & GNS & HOGN & HDGN & HHOGN & GNSTODE \\ \midrule
\multirow{6}{*}{Gravity}& \multirow{3}{*}{RMSE} & 0.083 & \underline{0.1971} & 0.5036  & 0.3124  & 0.5055 & \textbf{0.1932} \\ 
& & 0.42  & \underline{0.8654}  & 1.1053  & 0.8114  & 1.1210 & \textbf{0.6936} \\  
& & 1.63  & 1.5903  & 1.5109  & \underline{1.2862}  & 1.5668 & \textbf{1.2595} \\ \cline{2-8} 
& \multirow{3}{*}{\begin{tabular}[c]{@{}l@{}}Energy\\Error\end{tabular}} & 0.083 & 0.0094  & 0.1804  & \underline{0.0220}  & 0.1762 & \textbf{0.0090} \\  
& & 0.42  & 0.0658  & 0.4226  & \underline{0.0584}  & 0.4237 & \textbf{0.0149} \\ 
& & 1.63  & 0.5787  & 0.7136  & \underline{0.0484}  & 0.8233 & \textbf{0.0464} \\ \midrule
\multirow{6}{*}{Coulomb} & \multirow{3}{*}{RMSE} & 0.083 & 0.2139  & \underline{0.2049}  & 0.2083  & 0.2556 & \textbf{0.2044} \\ 
& & 0.42  & \underline{0.9040} & 1.4009  & 0.9442  & 1.3949 & \textbf{0.7468} \\  
& & 1.63  & 0.4660  & 0.4574 & \underline{0.4568} & 0.4851 & \textbf{0.2009} \\ \cline{2-8} 
& \multirow{3}{*}{\begin{tabular}[c]{@{}l@{}}Energy\\Error\end{tabular}} & 0.083 & 1.1377  & \textbf{0.7508}  & 0.9783  & 1.5492 & \underline{0.9576} \\ 
& & 0.42  & 0.0752 & 0.0933 & \underline{0.0499}  & 0.8810 & \textbf{0.0313} \\ 
& & 1.63  & 3.7563  & 3.7467  & \underline{3.6365}  & 4.7369 & \textbf{1.1669} \\ \bottomrule
\end{tabular}}
\label{table: intensity}
\end{table*}

\subsection{Baseline Methods}
We compare the proposed GNSTODE model with the following four competitive learning based particle system simulation baselines:

\begin{itemize} 

	\item \textbf{GNS}~\cite{sanchez2020learning} employs multi-step message passing to capture the complex particle-particle interactions. 
	
	\item \textbf{HOGN}~\cite{sanchezgonzalez2019hamiltonian} utilizes the GIN~\cite{battaglia2018relational} to model the Hamiltonian of particle systems and applies the Hamiltonian to capture the complex interactions between particles.
 


    \item \textbf{HDGN}~\cite{martinkus2021scalable} constructs a hierarchical graph to model particles' multi-level neighbors and leverages the GIN~\cite{battaglia2018relational} to capture the cross-level interactions.

	\item \textbf{HHOGN}~\cite{martinkus2021scalable} extends the HOGN model~\cite{sanchezgonzalez2019hamiltonian} to the hierarchical particle interaction graph for achieving more accurate particle system simulation.
 

\end{itemize}

\begin{table*}[t]
\scriptsize
\centering
\caption{Simulation performance comparison on the Gravity and Coulomb systems under the setting of varying time steps. The best and second best performers are respectively highlighted by the \textbf{boldface} and \underline{underline}.
}
\resizebox{1.8\columnwidth}{!}{
\begin{tabular}{l|l|c|c c c c c}
\toprule
Dataset & Metrics $\downarrow$ & Step & GNS & HOGN & HDGN & HHOGN & GNSTODE \\ \midrule
\multirow{6}{*}{Gravity} & \multirow{3}{*}{RMSE} & 5 & 0.9557 & 1.4083 & \underline{0.8817} & 1.4068 & \textbf{0.7517} \\ 
 & & 10 & 1.0739 & 1.5242 & \underline{1.0213} & 1.5292 & \textbf{0.7965}  \\  
 & & 20 & \underline{1.3126} & 1.6155 & 1.5292 & 1.7314 & \textbf{0.8833}  \\ \cline{2-8} 
 & \multirow{3}{*}{\begin{tabular}[c]{@{}l@{}}Energy\\ Error\end{tabular}} & 5 & \underline{0.0971} & 0.9396 & 0.9787 & 0.9835 & \textbf{0.0197} \\  
 & & 10 & 0.1571 & 1.5965 & \underline{0.1444} & 1.4450 & \textbf{0.0414}  \\ 
 & & 20 & 0.8050 & 1.7440 & \underline{0.2874} & 2.5928 & \textbf{0.0512}  \\ \midrule
 \multirow{6}{*}{Coulomb} & \multirow{3}{*}{RMSE} & 5 & \underline{0.9040} & 1.4006 & 0.9442 & 1.3949 & \textbf{0.7468}  \\  
 & & 10 & \underline{0.9903} & 1.5084 & 1.0656 & 1.5188 & \textbf{0.7436}  \\ 
 & & 20 & 1.1939 & 1.5937 & 1.2679 & 1.6505 & \textbf{0.8487}  \\ \cline{2-8} 
 & \multirow{3}{*}{\begin{tabular}[c]{@{}l@{}}Energy\\ Error\end{tabular}} & 5 & 0.0751 & 0.9325 & \underline{0.0849} & 0.8810 & \textbf{0.0313}  \\ 
 & & 10 & \underline{0.0663} & 1.4966 & 0.1284 & 1.4442 & \textbf{0.0225}  \\  
 & & 20 & \underline{0.2994} & 1.7826 & 0.3376 & 2.2266 & \textbf{0.0489}  \\ \bottomrule
\end{tabular}}
\label{table: L_steps}
\end{table*}

\subsection{Evaluation Metrics}
Following \cite{martinkus2021scalable}, we evaluate the particle system simulation performance using the following two metrics:
\begin{itemize}
    \item \textbf{RMSE (Root Mean Square Error)}. We expect the simulated particle trajectories to be as close to the ground truth as possible. RMSE measures the discrepancy between the predicted and ground-truth trajectories:
    \begin{equation}
        \text{RMSE} =\sqrt{ \frac{1}{N} \sum_{i=1}^{N}\sum_{t=2}^{T} \Vert X^c_{i,t} - \hat{X}^c_{i,t}\Vert_{\mathrm{F}}^2},
    \label{equ:rmse}
    \end{equation}where $N$ is the number of test particle system trajectories, $T$ is the number of time stamps of the test trajectories, and $X^c_{i,t}$ and $\hat{X}^c_{i,t}$ are respectively the ground-truth and predicted particle coordinates at the time stamp $t$.
    \item \textbf{Energy Error}. Both the Gravity and Coulomb systems follow the principle of energy conservation, meaning that the total energy of the whole system should be constant over time. The Energy Error measures the extent to which the simulation violates the principle of energy conservation:
    \begin{equation}
        \text{Energy Error} = \frac{1}{N} \sum_{i=1}^{N} \frac{H_{i,1} - \hat{H}_{i,T}}{H_{i, 1}},
    \label{equ:energy}
    \end{equation}where $H_{i,1}$ and $\hat{H}_{i,T}$ are respectively the Hamiltonian (total energy) of the ground-truth system state at time stamp $1$ and the predicted system state at time stamp $T$ of the $i$-th test trajectory. 
\end{itemize}

For both the RMSE and Energy Error metrics, a lower score indicates a better simulation performance. 





\subsection{Implementation Details}

All experiments are conducted on a personal computer with the Ubuntu 18.04 OS, an NVIDIA GeForce RTX 2080Ti (12GB memory) GPU, an Intel Core i9-10900X (3.70 GHz) CPU and a 32 GB RAM. Default parameter configurations are used for implementing baseline models. 

When implementing the GNSTODE model, 15-nearest neighbor graph is used to construct the spatial graph and the latent particle embeddings' dimension is set same as the original particle features' dimension (5 for the Gravity system and 6 for the Coulomb system). The maximum number of epochs is set to 200. At each epoch, we collect the paired particle system states at every two consecutive time stamp from the training trajectories, shuffle them and split them into a number of batches, then iterate each batch of particle state pairs to minimize the corresponding reconstruction errors. The batch size is set to 50. We optimize the GNSTODE model with the Adam optimizer~\cite{adam_kingma2014adam}.

For the proposed GNSTODE model and baseline models, the validation set is employed to select the best epoch, and the test set is used to evaluate the simulation performance.
On each dataset and under each setting, we run the proposed GNSTODE model and baseline models for five times with different random parameter initializations and report the averaged RMSE and Energy Error scores as models' simulation performance.

\subsection{Performance Comparison} \label{subsec:exp_ad_result} 

Tables \ref{table: scale}-\ref{table: L_steps} compare the simulation performance of different models under the three different settings, i.e., varying scales, varying intensities and varying time steps. For each comparison, the best and second best performers are respectively highlighted by the \textbf{boldface} and \underline{underline}. From Tables \ref{table: scale}-\ref{table: L_steps}, we can find that the proposed GNSTODE model consistently achieves the best RMSE and Energy Error scores on each dataset and under each setting, except for the second best Energy Error on the Couloub system with intensity 0.083.

Table \ref{table: scale} shows that the proposed GNSTODE model significantly outperforms the compared baseline methods on different scales of particle systems. This verifies that the proposed GNSTODE model is robust to the varying scales of particle systems. As is shown in Table \ref{table: scale}, models tend to achieve better simulation performance on the particle systems with larger scales, which provide more supervision on particle movement patterns. 

From Table \ref{table: intensity}, we can see that the proposed GNSTODE model achieves superior RMSE and Energy Error scores than baseline models on particle systems with varying intensities. This is mainly contributed by the effectiveness of the proposed neural spatial ODE component in capturing the varying spatial dependencies between particles. The slight performance drop on the particle systems with larger intensities indicates that the larger spatial dependency variations bring more difficulties to the learning based particle system simulation. 

Table \ref{table: L_steps} demonstrates the best simulation performance of the proposed GNSTODE model on the particle systems with different time steps. This is consistent with the fact that the designed neural temporal ODE makes the proposed GNSTODE able to adapt the varying temporal dependencies between every two consecutive time stamps. As large time steps incorporate more temporal dependency variations, long-time-step simulation usually yields higher RMSE and Energy Error scores.

\subsection{Simulation Visualization}

\begin{figure*}
\centering
\subfigure[Ground Truth]{
\includegraphics[width=.20\linewidth]{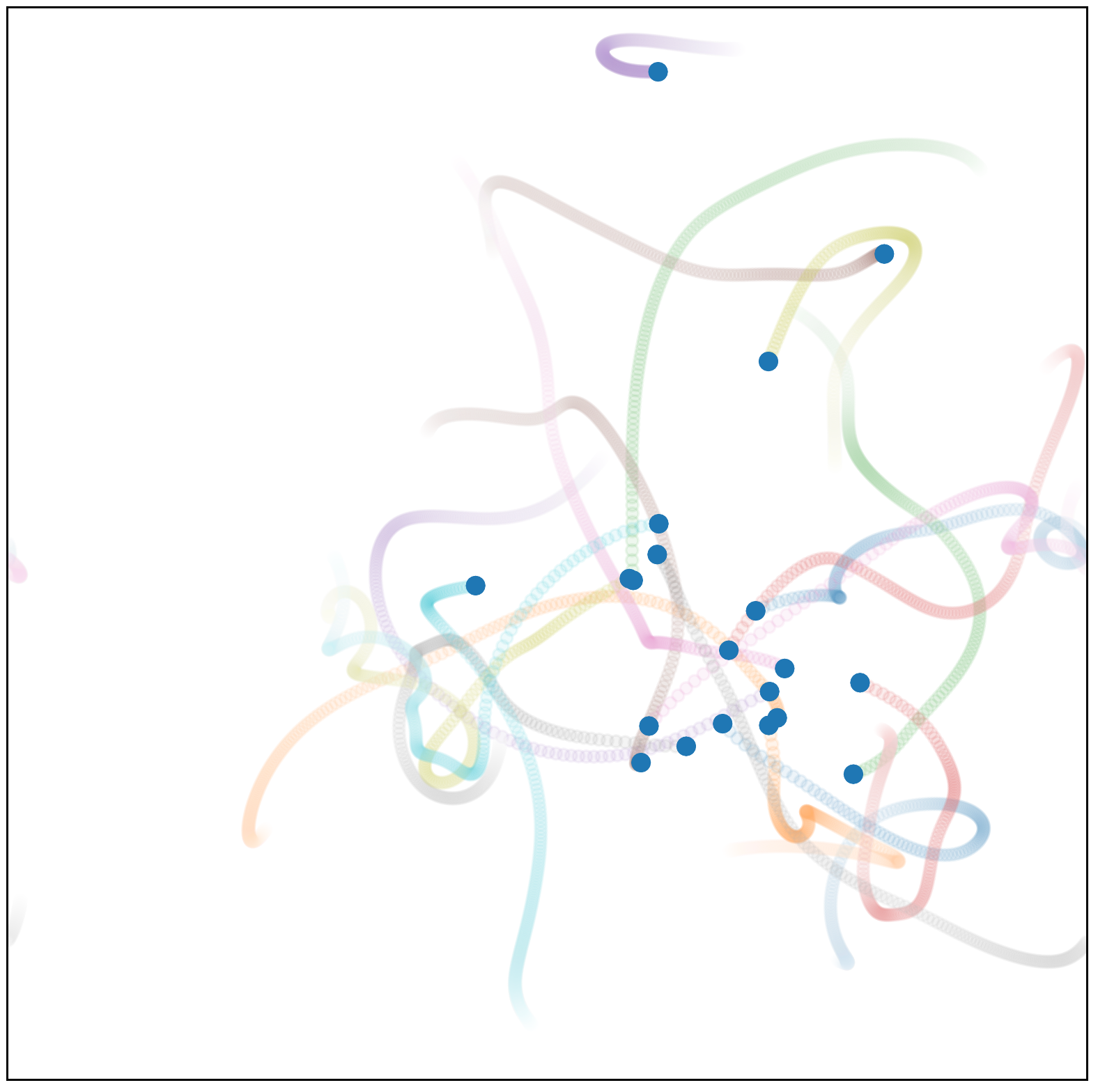}

\label{vfig: Target}
}
\subfigure[GNS]{
\includegraphics[width=.20\linewidth]{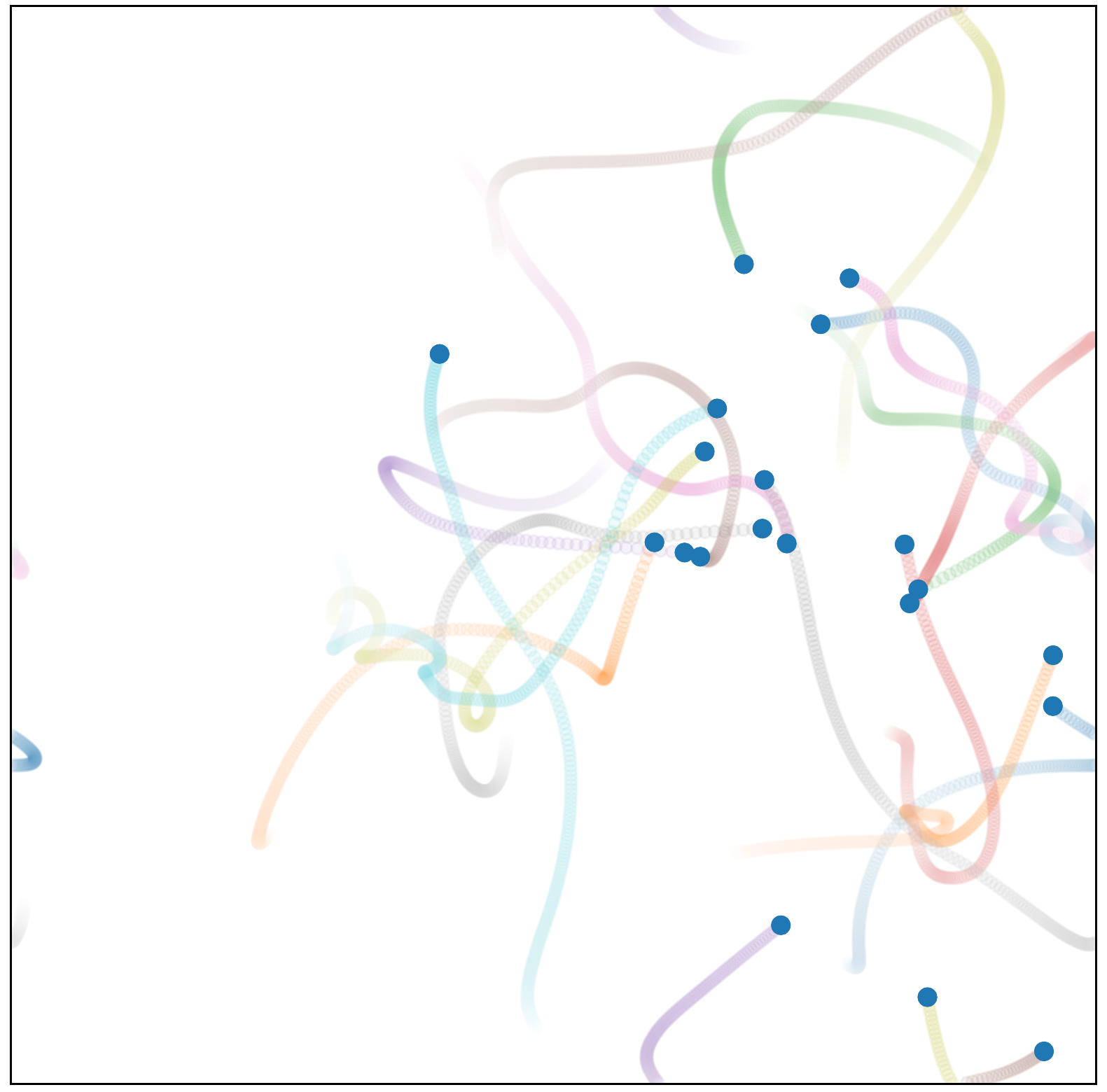}

\label{vfig: GN}
}
\subfigure[HDGN]{
\includegraphics[width=.20\linewidth]{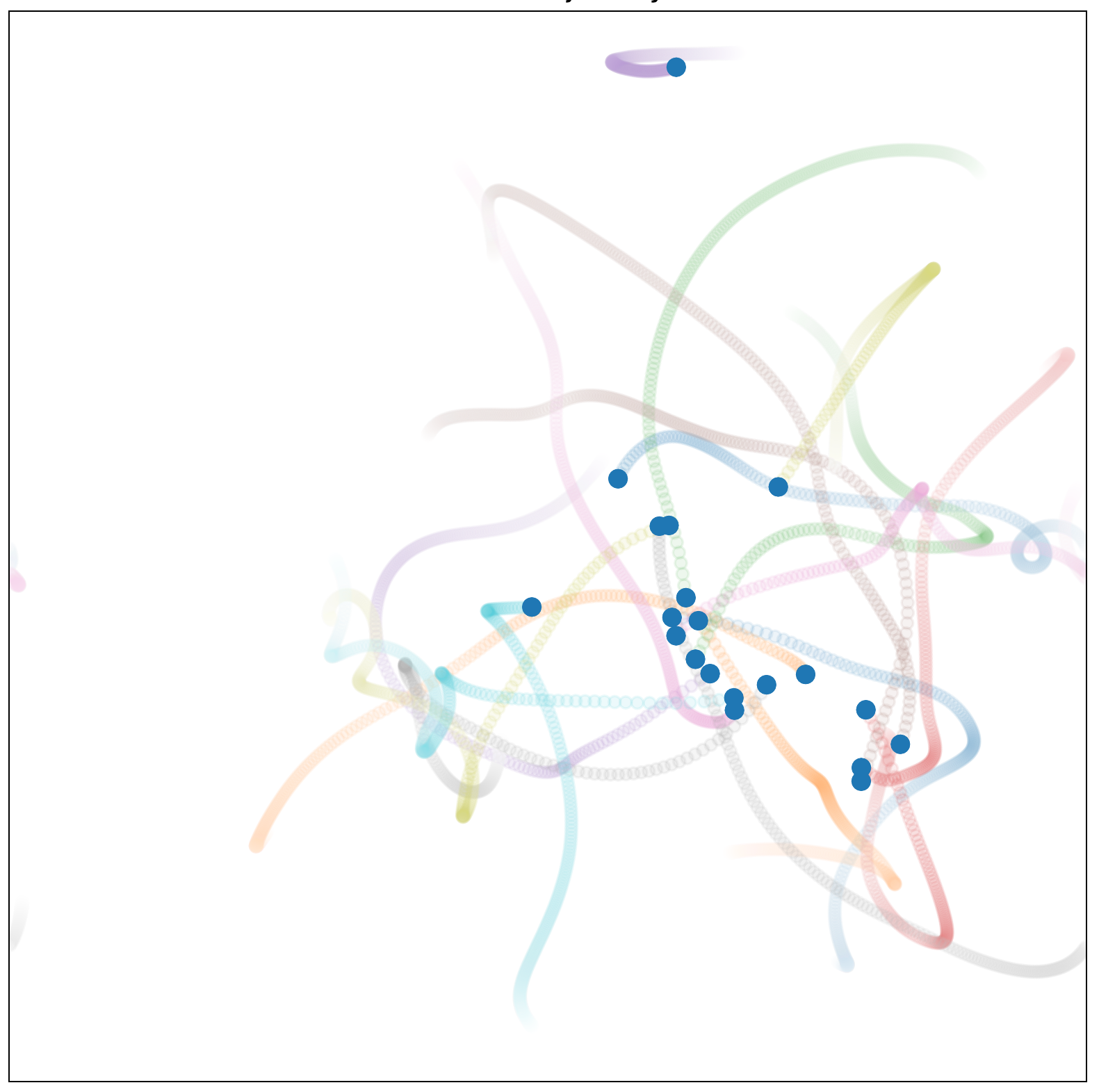}

\label{vfig: HDGN}
}
\subfigure[GNSTODE]{
\includegraphics[width=.20\linewidth]{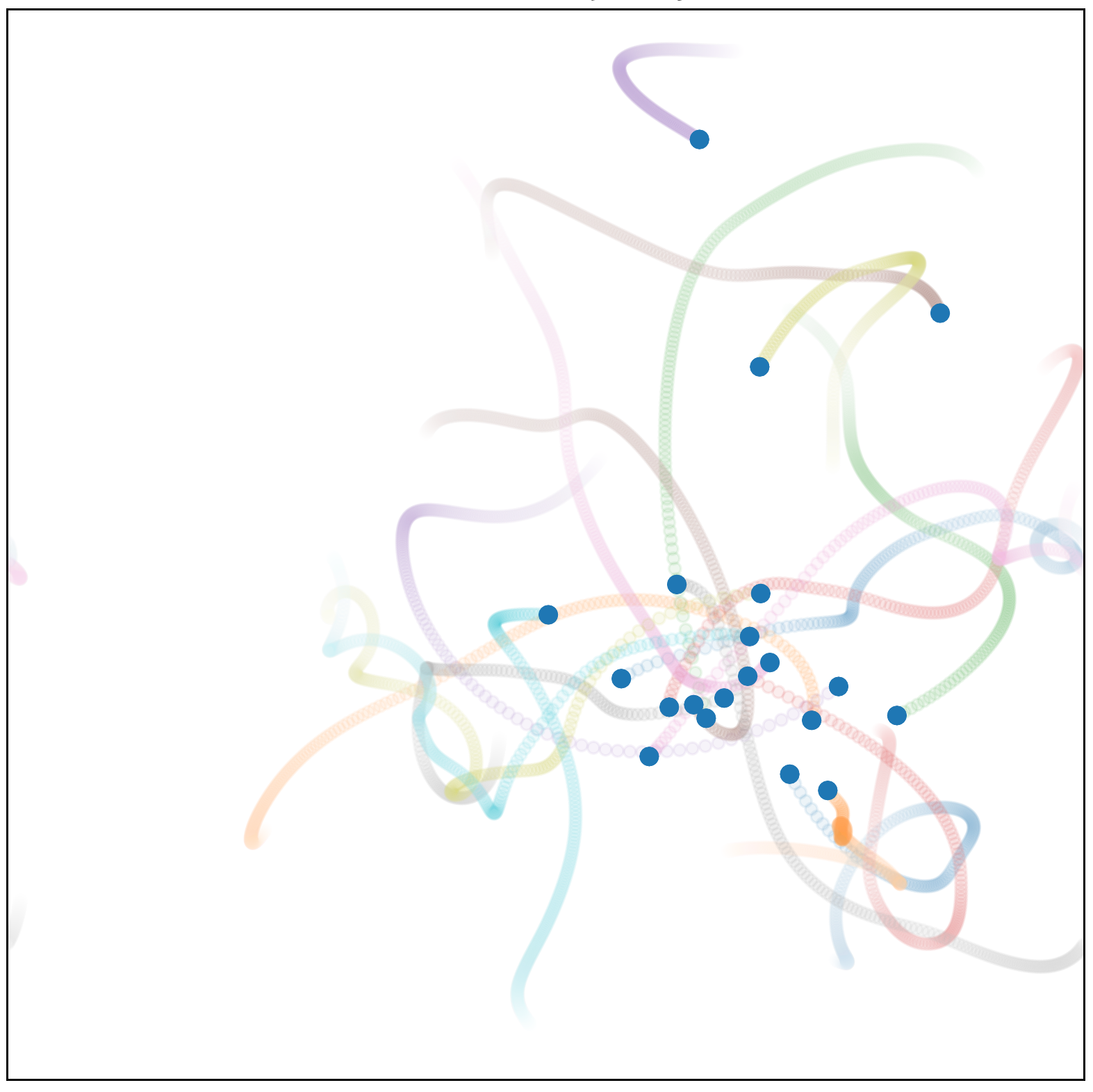}

\label{vfig: GNSTODE}
}

\caption{
The ground-truth trajectory and simulations of a Gravity system with scale 20, intensity 0.42 and time step 1.
}
\label{fig:appendix_20_0.42_1}
\end{figure*}

\begin{figure*}[t]
\centering
\subfigure[Ground Truth]{
\includegraphics[width=.20\linewidth]{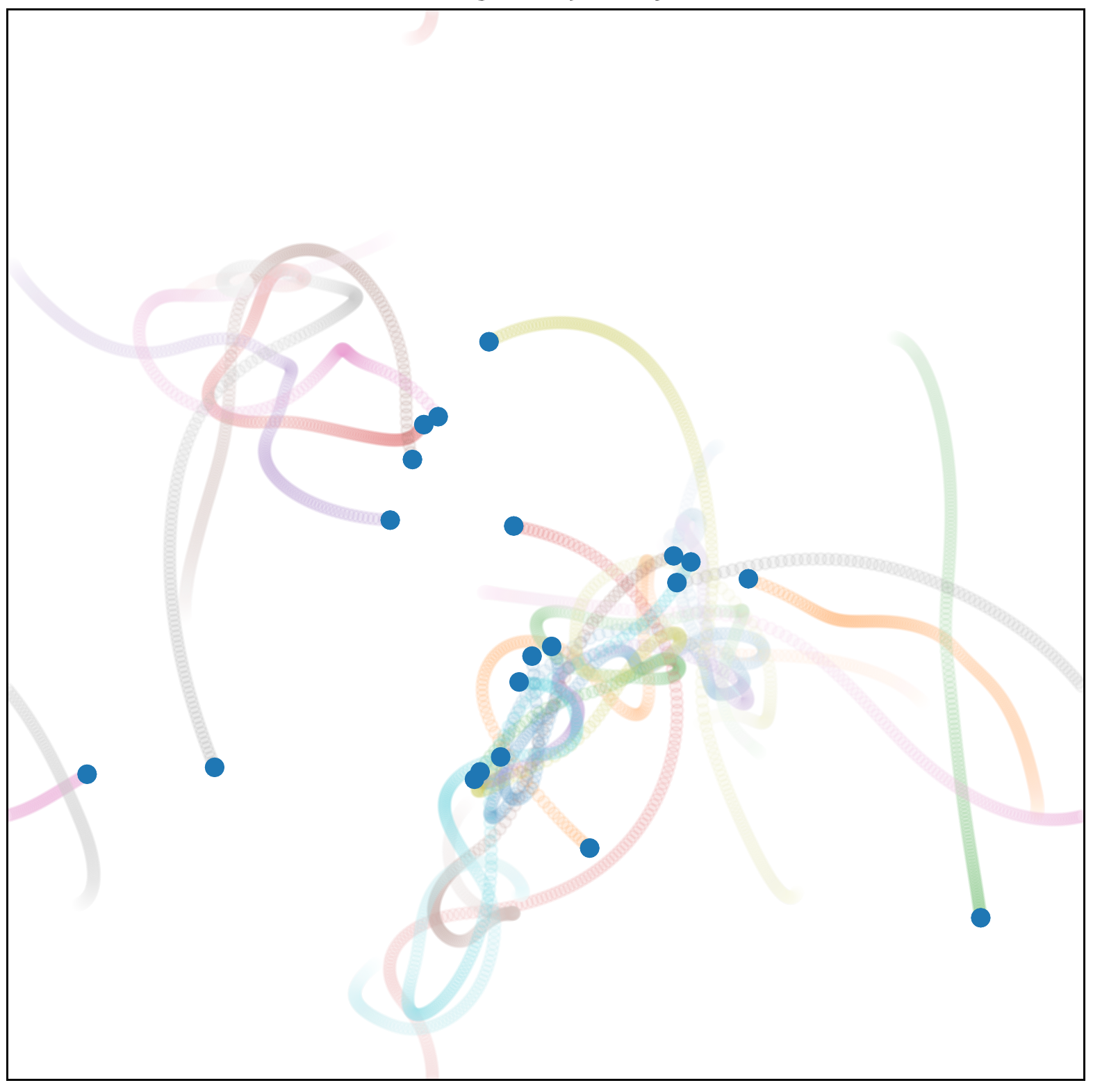}
\label{apx:target_c}
}
\subfigure[GNS]{
\includegraphics[width=.20\linewidth]{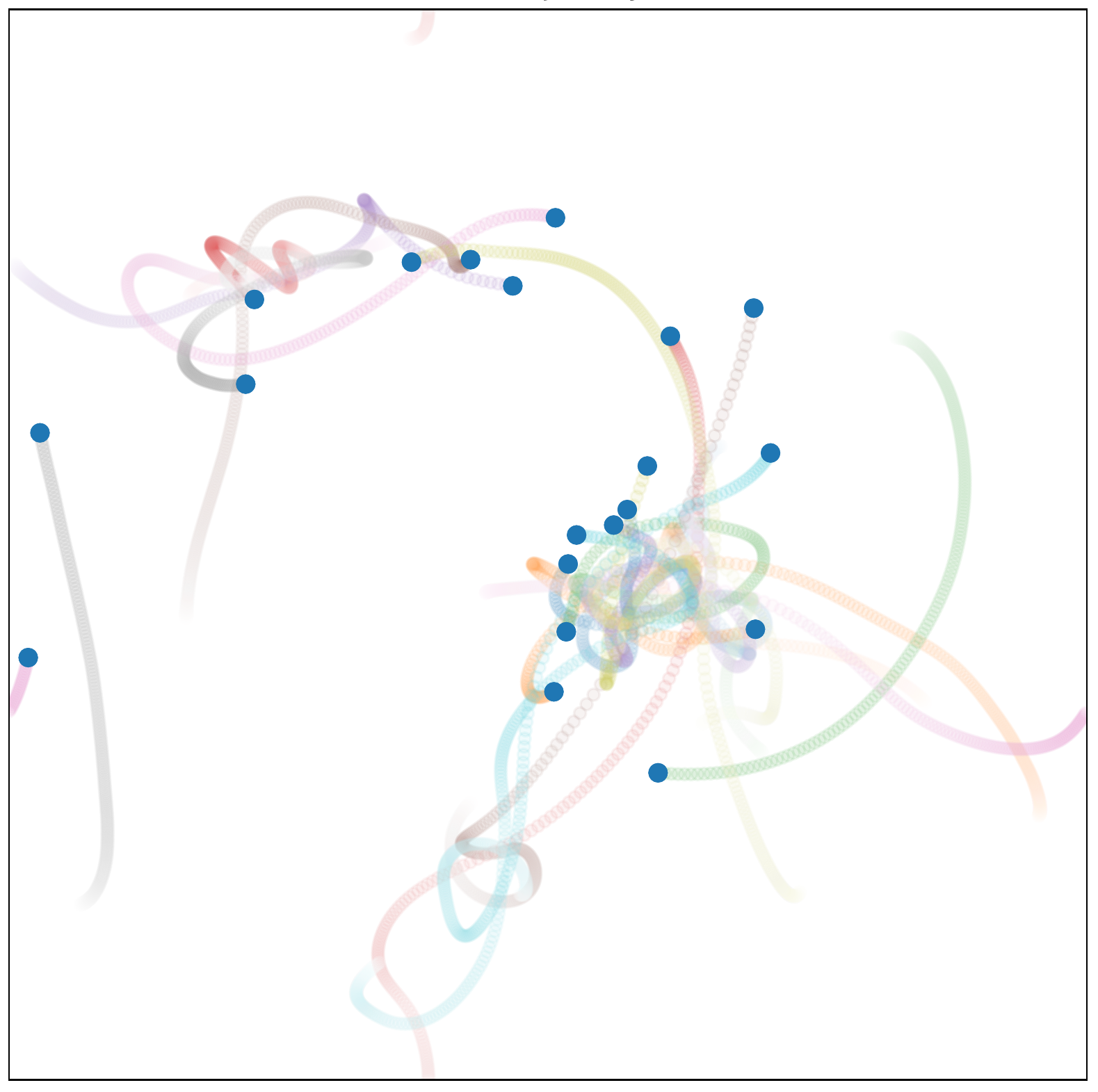}
\label{apx:GN_c}
}
\subfigure[HDGN]{
\includegraphics[width=.20\linewidth]{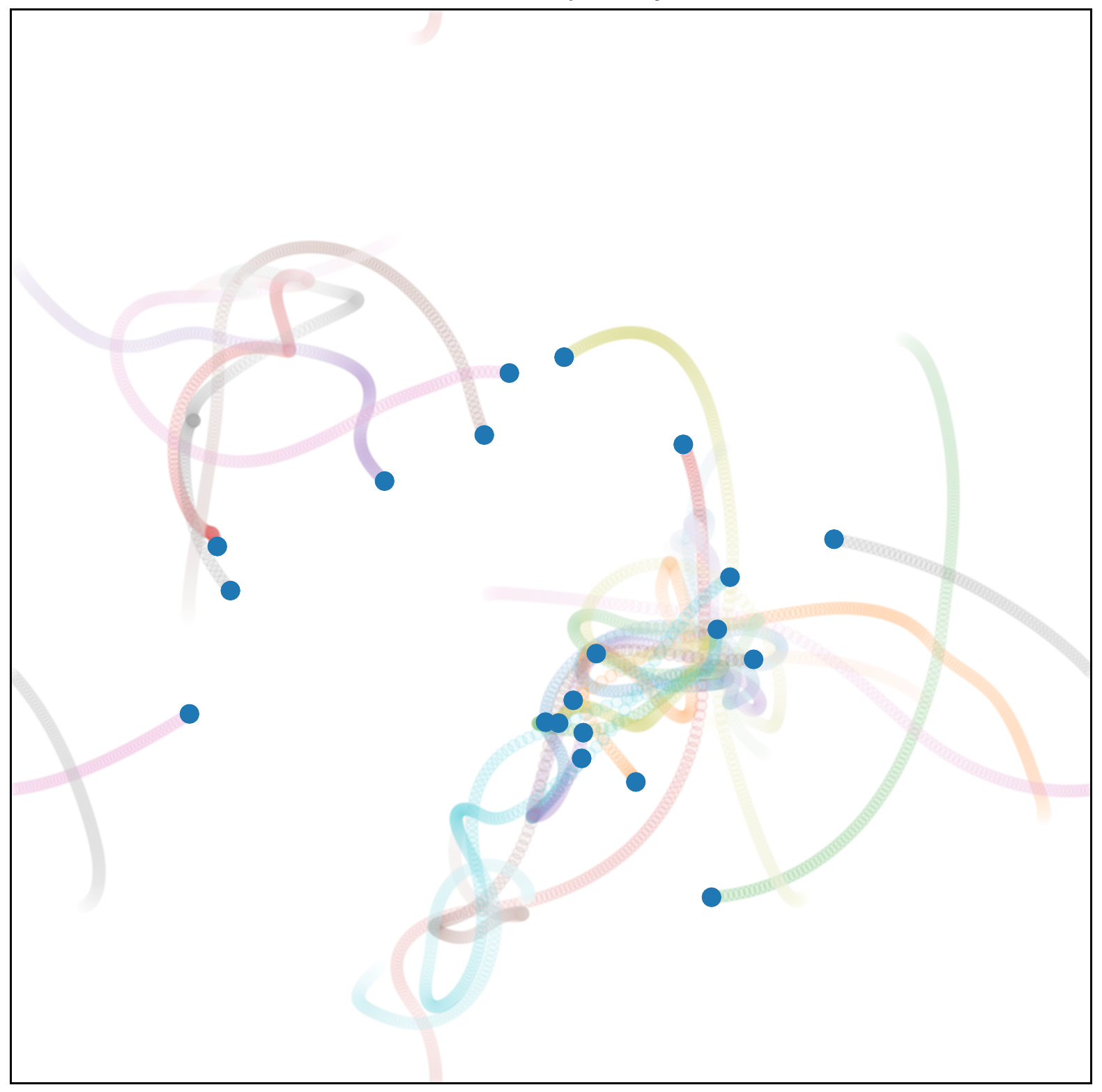}
\label{apx:HDGN_c}
}
\subfigure[GNSTODE]{
\includegraphics[width=.20\linewidth]{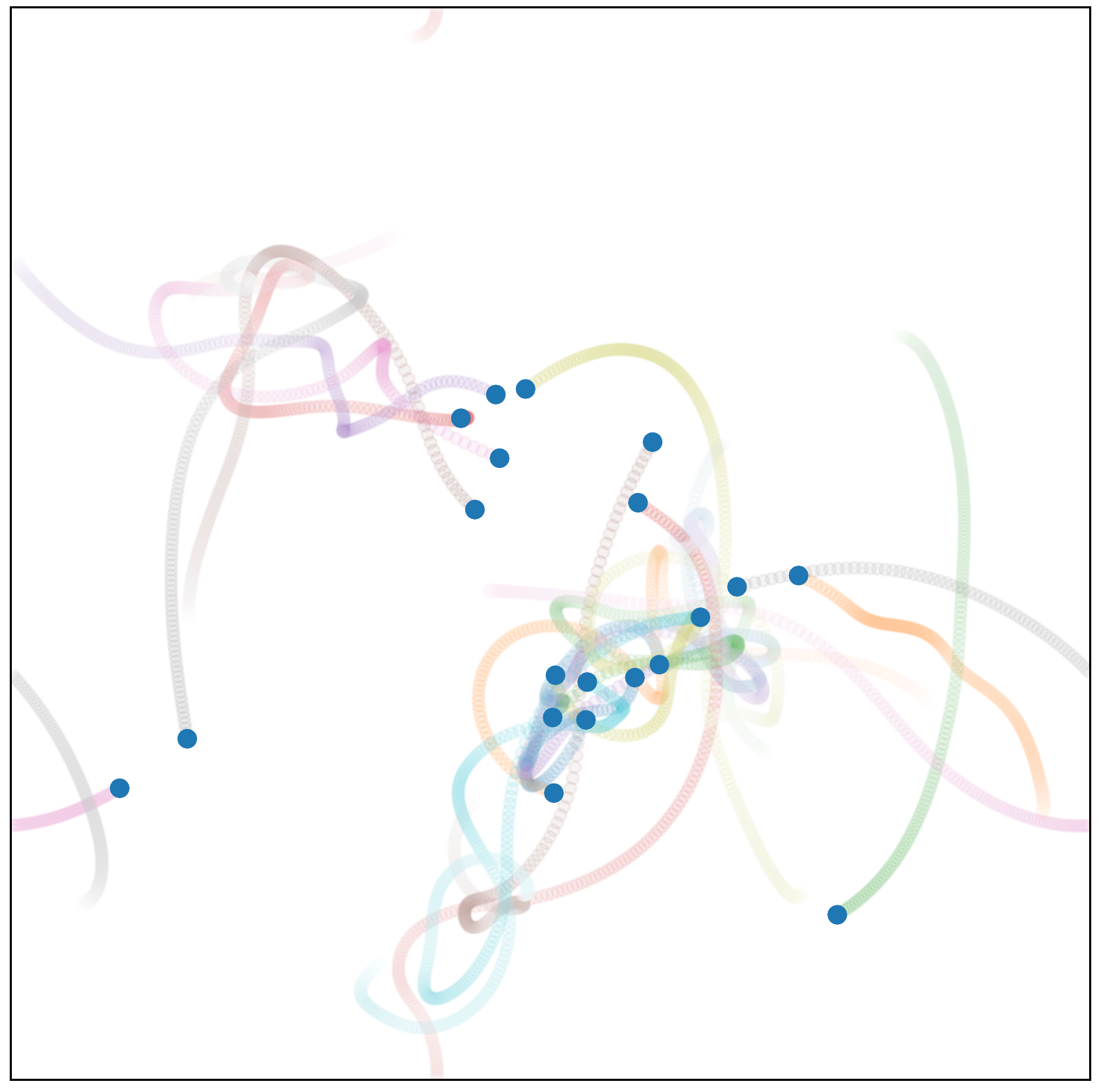}
\label{apx:GNSTODE_c}
}
\caption{The ground-truth trajectory and simulations of a Coulomb system with scale 20, intensity 0.42 and time step 1.}
\label{fig:appendix_c}
\end{figure*}

In Fig. \ref{fig:appendix_20_0.42_1} and \ref{fig:appendix_c}, we visualize the simulations produced by the GNS, HDGN and the proposed GNSTODE model on a Gravity system and a Coulomb system with scale 20 and intensity 0.42, and compare them with the ground-truth trajectories. Through the comparison, we can find that the simulations produced by the GNS and HDGN baseline models have significant discrepancies with the ground-truth trajectories, while the GNSTODE simulations can be almost perfectly aligned to the ground-truth trajectories. This further verifies the superior simulation performance of the proposed GNSTODE model over the baseline models. 

\subsection{Ablation Study} \label{subsec:exp_ablation_study}    

To further verify the effectiveness of the proposed neural spatial and temporal ODE components, we also conduct an ablation study by comparing the full GNSTODE model with its two ablated variants:
\begin{itemize}
    \item \textbf{w/o Spatial ODE}. For this variant, we replace the neural spatial ODE component in Eq. (\ref{spatial_ode_solution}) with the GIN message-passing operation in Eq.(\ref{GIN_skip_layer}) to model the interactions between particles. 
    \item \textbf{w/o Temporal ODE}. For this variant, we replace the neural temporal ODE component in Eq. (\ref{temporal_ode_solution}) with the simple addition operation in Eq. (\ref{naive_temporal_updating}) to predict the next time stamp's system state. 
\end{itemize}

Tables \ref{table: ablation study rmse}-\ref{table: ablation study energy} respectively compare the RMSE and Energy Error scores of the full GNSTODE model and its ablated variants on the Gravity system with different scale, intensity and time step configurations. The best performer is highlighted by the \textbf{boldface} for each comparison. From Tables \ref{table: ablation study rmse}-\ref{table: ablation study energy}, we can find that the full GNSTODE consistently outperforms its ablated variants -- w/o Spatial ODE and w/o Temporal ODE -- by large margins under all configurations. This proves that the neural spatial and temporal ODE components are both important to the high-quality GNSTODE simulations. Ablating either of them will result in inferior simulation performance.

\begin{table*}[t]
    \small
    \centering
 	\caption{
  The RMSE scores of the full GNSTODE model and its ablated variants on the Gravity system with different scale, intensity and time step configurations. The best performers are highlighted by the \textbf{boldface}.
   }
	\begin{tabular}{p{85 pt}<{} p{45 pt}<{\centering} p{45 pt}<{\centering} p{45 pt}<{\centering} p{45 pt}<{\centering} p{45 pt}<{\centering} p{45 pt}<{\centering}}
		
		\toprule
		    & \multicolumn{2}{c}{{Scale}} & \multicolumn{2}{c}{{Intensity}} & \multicolumn{2}{c}{{Time Step}} \\
          \cmidrule{2-7}
		  Method & 20  & 100 & 0.083 & 1.63 & 5 & 10  \\
		  
		  \midrule
		  
		  Full GNSTODE & \textbf{0.7053} & \textbf{0.9937} & \textbf{0.1934} & \textbf{1.2595} & \textbf{0.7351} & \textbf{0.7965}  \\
		  
		  \midrule
          w/o Spatial ODE & 0.8414 & 1.5549 & 0.2142 & 1.3644 & 0.8825 &0.9901\\
		  w/o Temporal ODE & 0.7437 & 1.0134 & 0.1954 & 1.2839 & 0.7422 & 0.8434\\
		\bottomrule
	
	\end{tabular}
	\label{table: ablation study rmse}
\end{table*}

\begin{table*}[t]
    \small
    \centering
 	\caption{
  The Energy Error scores of the full GNSTODE model and its ablated variants on the Gravity system with different scale, intensity and time step configurations. The best performers are highlighted by the \textbf{boldface}.
   }
	\begin{tabular}{p{85 pt}<{} p{45 pt}<{\centering} p{45 pt}<{\centering} p{45 pt}<{\centering} p{45 pt}<{\centering} p{45 pt}<{\centering} p{45 pt}<{\centering}}
		
		\toprule
		    & \multicolumn{2}{c}{{Scale}} & \multicolumn{2}{c}{{Intensity}} & \multicolumn{2}{c}{{Time Step}} \\
          \cmidrule{2-7}
		  Method & 20  & 100 & 0.083 & 1.63 & 5 & 10  \\
		  
		  \midrule
		  
		  Full GNSTODE & \textbf{0.0167} & \textbf{0.0199} & \textbf{0.0090} & \textbf{0.0464}& \textbf{0.0196} & \textbf{0.0414}  \\
		  
		  \midrule
		  w/o Spatial ODE  & 0.0405 & 0.0308 & 0.0096 & 0.0495 & 0.0210  & 0.1304\\
	      w/o Temporal ODE & 0.0446 & 0.1557 & 0.0117 & 0.0945 & 0.0378  & 0.0912 \\
	
		\bottomrule
	
	\end{tabular}
	\label{table: ablation study energy}
\end{table*}

\subsection{Parameter Sensitivity Study} \label{subsec:exp_param_study} 

\begin{figure*}
\centering
\subfigure{
\includegraphics[width=.3\linewidth]{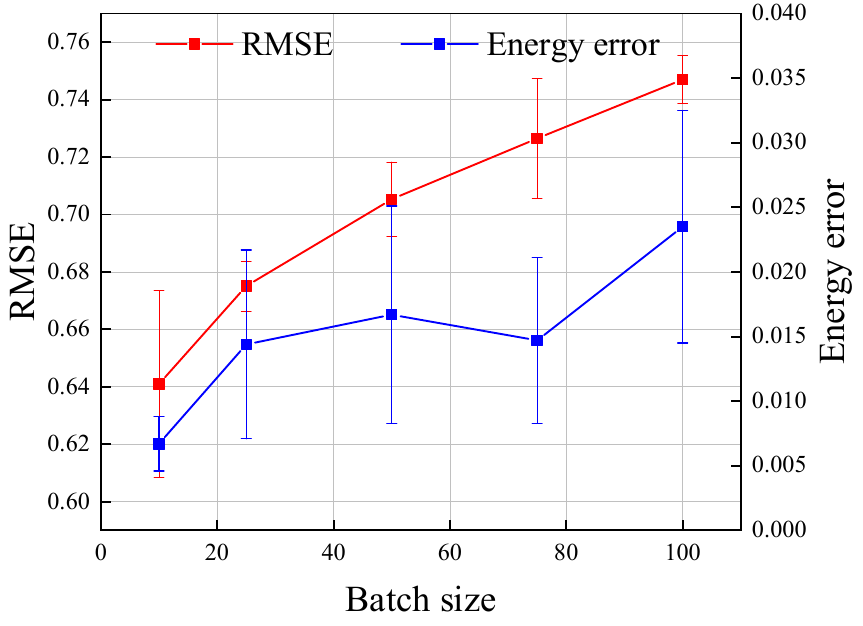}
\label{subfig: param_batch_size}
}
\subfigure{
\includegraphics[width=.3\linewidth]{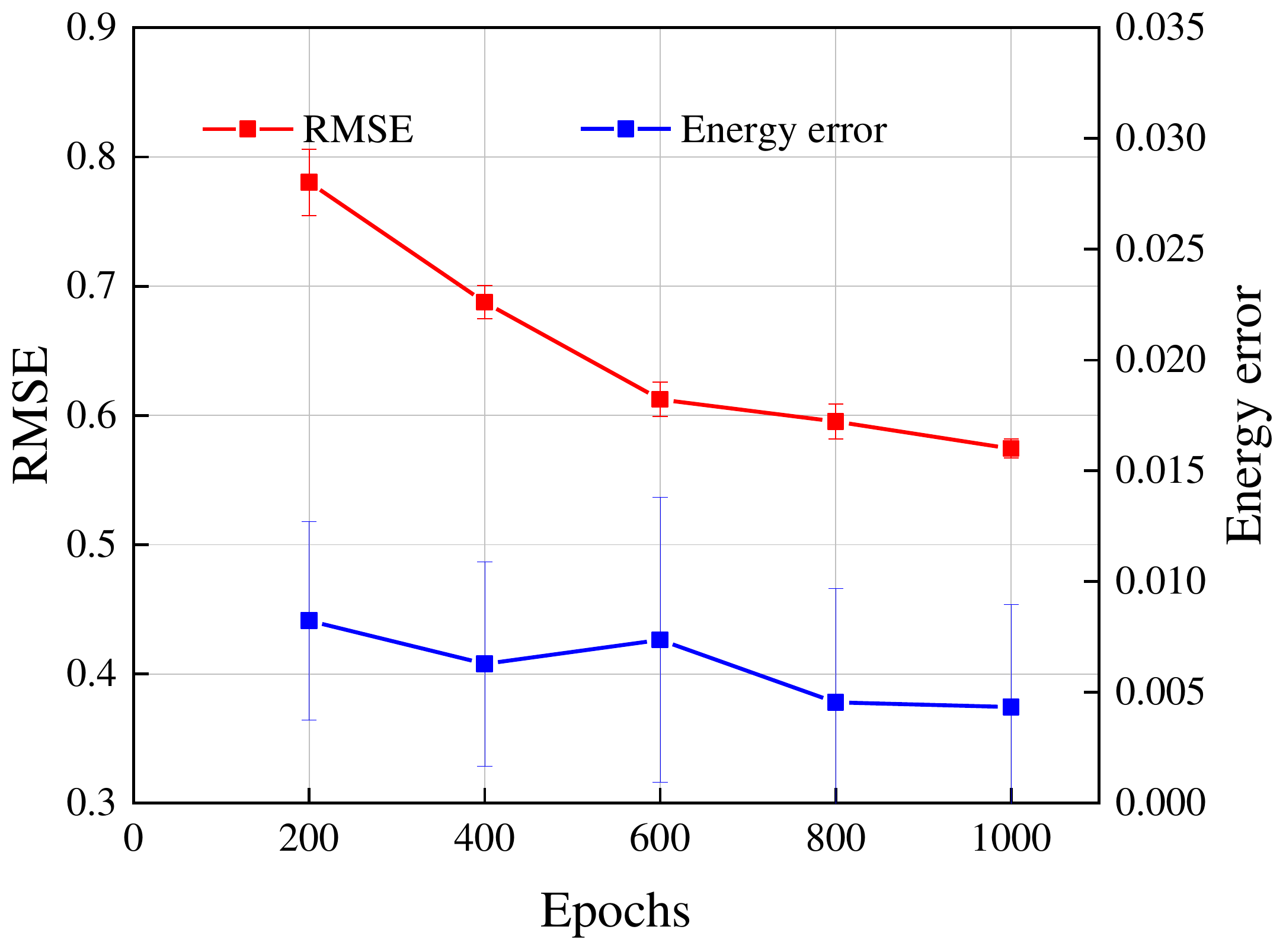}
\label{subfig: param_epoch}
}
\subfigure{
\includegraphics[width=.3\linewidth]{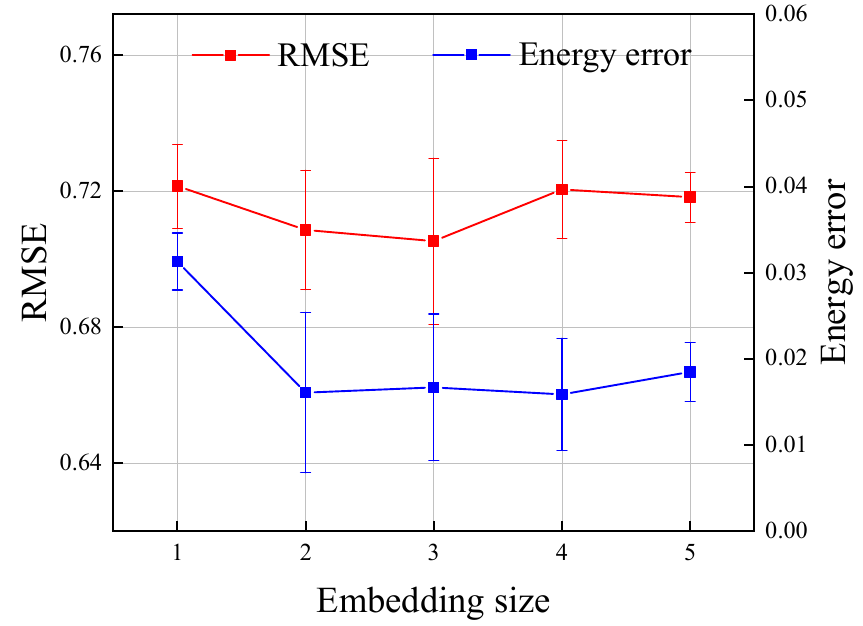}
\label{subfig: param_embed}
}
\caption{Parameter sensitivity with regard to the batch size, epoch number and embedding size.}
\label{fig:para}
\end{figure*}

In this section, we study the sensitivity of the proposed GNSTODE model with regard to the three main model parameters, the batch size, the maximum number of epochs (epoch number) and the latent embeddings' dimension, on the Gravity dataset with scale, intensity and time step respectively set to 20, 0.42 and 1. In turn, we fix any two of the three parameters as default values, and study the performance change of the GNSTODE model when the value of the remainder parameter varies in a proper range. Fig.~\ref{fig:para} shows the model's sensitivity with regard to the three parameters. 


From Fig.~\ref{subfig: param_batch_size}, we can find that the model's performance degrades with the increase of batch size. The large batch size makes the gradient descent hard to capture the uniqueness of each particle system's updating with varying spatial and temporal dependencies, resulting in inferior simulation performance. Fig.~\ref{subfig: param_epoch} shows that the GNSTODE's performance gradually increases and then remains at a stable level with the increase of the epoch number. The study indicates that the epoch number is required to be large enough to make sure the model is sufficiently trained, while an overlarge epoch number is not necessary when the model training is saturated. Fig.~\ref{subfig: param_embed} demonstrates that the GNSTODE's performance is not sensitive to the dimension of particle latent embeddings. This is mainly contributed by the strong learning ability of the proposed neural spatial and temporal ODE components for modeling the complicated particle system state updates, making GNSTODE still able to achieve satisfactory performance even with a small particle embedding dimension.

\section{Conclusion and Future Work}\label{sec:conclusion}
In this paper, we employed the GNNs and dual neural ODEs in the both spatial and temporal domains to tackle the problem of learning to simulate particle systems for the first time, established a new paradigm for physics learning and outlined a comprehensive solution procedure. Moreover, to verify the efficacy of our model, we designed three experimental settings on the Gravity and Coulomb particle systems, i.e., varying scales, varying intensities and varying time steps, which will also inspire the follow-up research to evaluate the learning based particle system simulation models comprehensively. The experimental results show that our model consistently achieves better simulation performance than the state-of-the-art learning based simulation models in all cases. More importantly, the results confirm our model's robustness to varying scales of particle systems and the ability to automatically adapt to particle systems with varying spatial and temporal dependencies. 



The changes in the actual physical world are significantly more complex than we believe. Although our model provides a solution to modeling the complex spatial and temporal dependencies in particle systems, it still has a long way to go before we can fully comprehend the laws of the real world. In addition to addressing the studied simulation problem, we can make further explorations by studying a series of related tasks, such as modeling the environment interaction (i.e., modeling the interactions between particles and boundaries) and modeling the interactions between particles with different material compositions, which are great value to process engineering. Another interesting future work is to study how to effectively inject physical inductive bias into the learning based simulation models to make them more consistent with the conventional Euler and Lagrangian simulation methods~\cite{subramaniam2013lagrangian} in rationale. Overall, the research on AI for physics science needs to be further explored thoroughly.  


\bibliography{reference}
\bibliographystyle{IEEEtran}

\end{document}